\documentclass{svjour3}                     
\smartqed  
\usepackage{amssymb}
\usepackage{amsmath}
\usepackage{latexsym}
\usepackage{graphicx}
\usepackage{mathptmx}      
\usepackage{bm}           
\usepackage{framed,color}
\definecolor{shadecolor}{rgb}{0.9,0.9,0.9}
\usepackage{rotating}
\usepackage{multirow}

\newcommand{\R}{\,{\mathbb R}}
\newcommand{\C}{\,{\mathbb C}}

\newcommand{\T}{\,{\mathbb T}}
\newcommand{\diag}{\mbox{diag}}

\newcommand{\cov}{\mbox{cov}}
\newcommand{\bolda}{{\mathbf a}}
\newcommand{\boldd}{{\mathbf d}}
\newcommand{\bolddhat}{\mbox{\boldmath$\hat\boldd$}}
\newcommand{\bolddtilde}{\mbox{\boldmath$\tilde\boldd$}}

\newcommand{\boldf}{{\mathbf f}}
\newcommand{\boldg}{{\mathbf g}}

\newcommand{\bolds}{{\mathbf s}}
\newcommand{\boldstilde}{\mbox{\boldmath$\tilde\bolds$}}

\newcommand{\boldx}{{\mathbf x}}
\newcommand{\boldy}{{\mathbf y}}
\newcommand{\boldA}{{\mathbf A}}

\newcommand{\boldF}{{\mathbf F}}
\newcommand{\boldG}{{\mathbf G}}
\newcommand{\boldH}{{\mathbf H}}

\newcommand{\calM}{\mathcal{M}}

\newcommand{\calN}{\mathcal{N}}
\newcommand{\boldP}{{\mathbf P}}
\newcommand{\boldQ}{{\mathbf Q}}

\newcommand{\boldTheta}{\boldsymbol{\Theta}}

\newcommand{\boldphi}{\boldsymbol{\varphi}}
\newcommand{\boldpsi}{\boldsymbol{\psi}}

\newcommand{\<}{\langle}
\renewcommand{\>}{\rangle}

\journalname{Circuits, Systems and Signal Processing}
\begin{document}

\title{Design of a Simple Orthogonal Multiwavelet Filter by Matrix Spectral Factorization%
}

\titlerunning{Design of a Simple Orthogonal Multiwavelet}        

\author{Vasil Kolev \and Todor Cooklev \and Fritz~Keinert}


\institute{Vasil Kolev \at
  Institute of Information and Communication Technologies \\
  Bulgarian Academy of Sciences \\
  Bl. 2 Acad. G. Bonchev St. \\
  1113 Sofia, Bulgaria \\
  \email{kolev\_acad@abv.bg} \\
  \and
  Todor Cooklev \at
  Wireless Technology Center \\
  Purdue University \\
  Fort Wayne, IN 46805, USA \\
  \email{cooklevt@pfw.edu} \\
  \and
  Fritz Keinert \at
  Dept.~of Mathematics \\
  Iowa State University \\
  Ames, IA 50011, USA \\
  \email{keinert@iastate.edu}
}

\date{Received: date / Accepted: date}

\maketitle

\begin{abstract}
  We consider the design of an orthogonal symmetric/antisymmetric
  multiwavelet from its matrix product filter by matrix spectral
  factorization (MSF). As a test problem, we construct a simple matrix
  product filter with desirable properties, and factor it using
  Bauer's method, which in this case can be done in closed form. The
  corresponding orthogonal multiwavelet function is derived using
  algebraic techniques which allow symmetry to be considered.  This
  leads to the known orthogonal multiwavelet SA1, which can also be
  derived directly.  We also give a lifting scheme for SA1,
  investigate the influence of the number of significant digits in the
  calculations, and show some numerical experiments.
  
\end{abstract}

\section{Introduction and Problem Formulation}
\label{sec:intro}

A digital filter bank is a collection of $r$ filters $\boldF_k(z)$, $k
= 0,\ldots,r-1$, which split a discrete-time signal into $r$
subbands. These subband signals are usually decimated by a factor of
$r$. Such a system is called a {\em maximally decimated analysis
  bank}. It can be used in applications such as signal, image, and
video coding. For example, a scalar wavelet system is based on a
single scaling function and mother wavelet, so $r=2$ \cite{SN-96}.  In
multiwavelet theory there are several scaling functions and mother
wavelets~\cite{CNKS-96}. This provides more degrees of freedom in the
design, and makes it possible to simultaneously achieve several useful
properties, such as symmetry, orthogonality, short support, and a
higher number of vanishing moments~\cite{SS-94a}.

The first multiscaling function (GHM) was constructed in 1994 by
Geronimo, Hardin, and Massopust~\cite{GHM-94}, based on fractal
interpolation. The multiwavelet function with which the GHM
multiscaling function forms a perfect reconstruction pair was
constructed in a separate paper by Strang and Strela~\cite{SS-95}.

Cooklev et al.~\cite{CNKS-96} investigated the general theory of $2
\times 2$ multifilter banks, and showed that real orthonormal
multifilter banks and multiwavelets can be obtained from orthonormal
complex-coefficient scalar filter banks. This ensures a number of good
properties, such as orthogonality, regularity, symmetry, and a
lattice-structure implementation. However, the limit functions in this
case tend to have long support.

The design of multifilter banks and multiwavelets remains a
significant problem. In the scalar case, spectral factorization of a
half-band filter that is positive definite on the unit circle $\T$ in
the complex plane was, in fact, the first design technique for perfect
reconstruction filter banks, suggested by Smith and
Barnwell~\cite{SB-86}. This provides motivation to extend Cooklev's
original idea~\cite{C95} to the multiwavelet case, where spectral
factorization is more challenging.

We consider a new approach for obtaining an orthogonal multiwavelet
which possesses symmetry/antisymmetry and regularity, from a matrix
product filter. This immediately leads to three main problems:

\begin{enumerate}
  \renewcommand{\labelenumi}{(\Alph{enumi})}
\item How can we obtain an appropriate matrix product filter which
    satisfies strong requirements?
  \item How do we perform Matrix Spectral Factorization (MSF)?
  \item	How do we improve the accuracy of the computed factors?
\end{enumerate}

We will consider these three problems in more detail, as well as
giving a brief description of some basic results.

Based on the previous theory and many applications of the SA1
multiwavelet, it is useful to develop a lifting scheme for it. Many of
the above applications can then be implemented in software or hardware
to be faster and more applicable.

The paper is organized as follows. After a brief review of
multiwavelet background, we begin by constructing a
half-band product filter with good smoothness properties. We then find
a multiscaling function by using matrix spectral factorization. To
complete the design by finding a multiwavelet function, we use additional
algebraic techniques based on QR decomposition. The resulting
multiwavelet is called SA1. It has orthogonality,
symmetry/antisymmetry, piecewise smooth functions, and short
support. We then construct a lifting scheme. Finally, some experiments
are shown with respect to this simple multiwavelet. This includes
investigating the influence of the number of significant digits used
in the calculations, as well as the performance of the SA1
multiwavelet in terms of coding gain.

The main novel contributions of this paper are the following:

\begin{enumerate}
  \item We introduce for the first time the product multifilter
    approach for designing a general MIMO filter bank;
  \item We obtain for the first time an orthogonal multiscaling
    function by using the closed form of the matrix spectral
    factorization algorithm of Youla and Kazanjian \cite{YK-78};
  \item We obtain the complementary orthogonal multiwavelet function;
  \item We construct a lifting scheme for the obtained simple
    orthogonal multiwavelet;
  \item We investigate the influence of the number of significant
    digits used for the coefficient $\sqrt{3}$, in 1D and 2D signal
    processing.
\end{enumerate}

\section{Multiwavelet Theory}
\label{sec:mwtheory}

We use the following notation conventions, illustrated with the letter 'a': 
\begin{itemize}
  \item[$a$] -- lowercase letters refer to scalars or scalar functions; 
  \item[$\bolda$] -- lowercase bold letters are vectors or vector-valued functions; 
  \item[$A$] -- uppercase letters are matrices; 
  \item[$\boldA$] -- uppercase bold letters are matrix-valued functions. 
\end{itemize}
$I$ and $0$ are identity and zero matrices of appropriate size. $A^T$
denotes the transpose of the matrix $A$, and $A^*$ is the complex
conjugate transpose. We are only considering real matrices in this
paper. However, the variable $z$ used in matrix polynomials lies on
the unit circle $\T$ in the complex plane, so that $\overline z =
z^{-1}$. Thus, the complex conjugate transpose of
\begin{equation*}
  \boldA(z) = \sum_k A_k z^k
\end{equation*}
is given by
\begin{equation*}
  \boldA(z)^* = \sum_k A_k^T z^{-k}. 
\end{equation*}
The inner product of two vector-valued functions $\boldf(t)$ and
$\boldg(t)$ in the $r$-dimensional Hilbert space $L^2(\R)^r$ of square
integrable functions is given by
\begin{equation*}
  \< \boldf, \boldg \> = \int \boldf(t) \boldg(t)^* \,dt,
\end{equation*}
which is a matrix.


\subsection{Multiwavelets and the Discrete Multiwavelet Transform}

{\em Multiscaling} and {\em multiwavelet functions} are vectors of $r$
scaling functions $\boldphi(t)$ and $r$ wavelet functions
$\boldpsi(t)$, respectively:
\begin{equation*}
  \boldphi(t) =
  \begin{bmatrix}
    \varphi_0(t) \\ \varphi_1(t) \\ \vdots \\ \varphi_{r-1}(t)
  \end{bmatrix}, \qquad 
  \boldpsi(t) =
  \begin{bmatrix}
    \psi_0(t) \\ \psi_1(t) \\ \vdots \\ \psi_{r-1}(t)
  \end{bmatrix}.
\end{equation*}
They belong to $L^2(\R)^r$, and satisfy the matrix dilation
equations~\cite{SN-96}
\begin{equation}\label{eq:01}
   \begin{split}
     \boldphi(t) &= \sqrt{2} \sum_{k=0}^m H_k \boldphi(2t-k), \\
     \boldpsi(t) &= \sqrt{2} \sum_{k=0}^m G_k \boldphi(2t-k).
   \end{split}
\end{equation}
Multiscaling and multiwavelet functions together are referred to as a
{\em multiwavelet}. All of the functions have support contained in the
interval $[0,m]$ (see \cite{WJ-97}).

We only consider orthonormal multiwavelets, where for any integer $\ell$, 
\begin{gather*}
  \< \boldphi(t), \boldphi(t - \ell) \> =  \< \boldpsi(t), \boldpsi(t - \ell) \> = \delta_{0\ell} I, \\
  \< \boldphi(t), \boldpsi(t - \ell) \> = 0.
\end{gather*}
Equivalently, the matrix coefficients $H_k$ and $G_k$ satisfy the
orthonormality conditions
\begin{equation}\label{eq:02}
   \begin{gathered}
     \sum_k H_k H_{k+2\ell}^T = \sum_k G_k G_{k+2\ell}^T = \delta_{0\ell} I, \\
     \sum_k H_k G_{k+2\ell}^T = 0.
   \end{gathered}
\end{equation}

In the {\em Discrete Multiwavelet Transform} (DMWT), the input signal
$\bolds^0$, at level $j=0$, is a sequence of $r$-vectors in
$\ell_2$. For example, in the 2-channel case,
\begin{equation*}
  \bolds^0 = \{ \bolds_k^0 \}_{k=-\infty}^\infty, \quad
  \bolds_k^0 =
  \begin{bmatrix}
    s_0^0[k] \\[5pt]
    s_1^0[k]
  \end{bmatrix}, \quad
  \mbox{with $\sum_k \|\bolds_k^0\|^2 < \infty$.}
\end{equation*}

The DMWT of $\bolds^0$ consists of computing recursively, for levels $j
= 1, \ldots J$, the approximation signal
\begin{equation}\label{eq:mwt1}
  \bolds^j = \{ \bolds_\ell^j \}_{\ell=-\infty}^\infty, \quad
  \bolds_\ell^j = \sum_k H_{k-2\ell} \bolds_k^{j-1}
\end{equation}
and the detail signal
\begin{equation}\label{eq:mwt2}
  \boldd^j = \{ \boldd_\ell^j \}_{\ell=-\infty}^\infty, \quad
  \boldd_\ell^j = \sum_k G_{k-2\ell} \bolds_k^{j-1}.
\end{equation}
We call $\bolds_\ell^j$ the {\em scaling function coefficients}, and
$\boldd_\ell^j$ the {\em wavelet coefficients}.

Because of orthogonality, the inverse can be computed iteratively in a
straightforward way:
\begin{equation}\label{eq:mwt3}
  \bolds_k^{j-1} = \sum_\ell H_{k-2\ell}^T \bolds_\ell^j + \sum_\ell
  G_{k-2\ell}^T \boldd_\ell^j.
\end{equation}
The matrix filter coefficients $H_k$, $G_k$ completely characterize
the multiwavelet filter bank.

We can also describe the filters and the wavelet coefficients of the
signal by their {\em symbols}
\begin{equation*}
  \boldH(z) = \sum_k H_k z^{-k}, \qquad
  \boldG(z) = \sum_k G_k z^{-k}, \qquad
  \bolds^j(z) = \sum_k \bolds_k^j z^k, \qquad
  \boldd^j(z) = \sum_k \boldd_k^j z^k.
\end{equation*}
In symbol notation, one step of DMWT and inverse DMWT is described by
\begin{equation}\label{eq:mwt4}
  \begin{split}
    \bolds^j(z^2) &= \frac{1}{2} \left[ \boldH(z) \bolds^{j-1}(z)  +
      \boldH(-z) \bolds^{j-1}(-z) \right], \\
    \boldd^j(z^2) &= \frac{1}{2} \left[ \boldG(z) \bolds^{j-1}(z)  +
      \boldG(-z) \bolds^{j-1}(-z) \right], \\
    \bolds^{j-1}(z) &= \boldH(z)^* \bolds^j(z^2) + \boldG(z)^* \boldd^j(z^2).
  \end{split}
\end{equation}

\begin{figure}
  \includegraphics[width=4.5in]{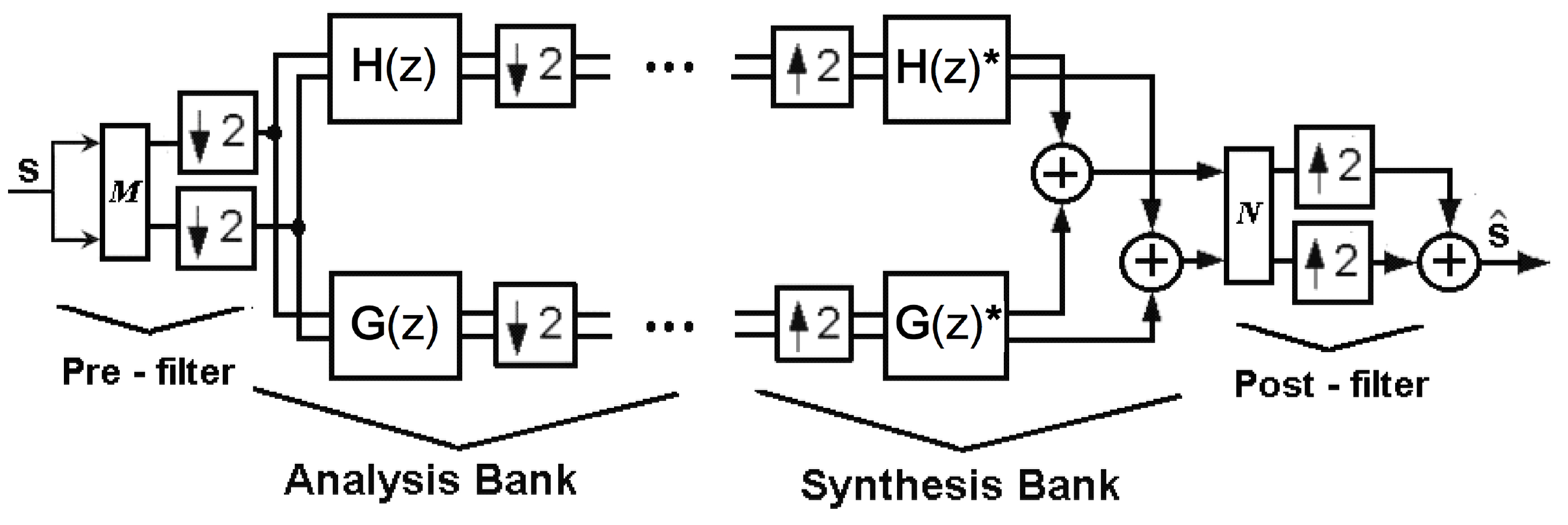}
  \caption{Two-channel critically sampled analysis and synthesis
    filter banks, with prefilter $\calM$ and postfilter $\calN$
    ($\downarrow 2$ and $\uparrow 2$ represent decimation and
    zero-padding).}
  \label{fig:01}
\end{figure}

Decomposition and reconstruction of the DMWT for a scalar input signal
$s$ is represented in fig.~\ref{fig:01}.  Since $s$ is
scalar-valued, it is necessary to vectorize it to produce $\bolds^0$.
A {\em prefiltering} step $\calM(z)$ is inserted at the beginning of
the analysis filter bank. A corresponding {\em postfilter} $\calN(z)$
is incorporated at the end, which is the inverse of the prefiltering
step.

The result of executing the steps in eq.~\eqref{eq:mwt4} is
\begin{equation*}
  \hat\bolds^0(z) = \frac{1}{2} \left[
    \boldH(z)^* \boldH(z) + \boldG(z)^* \boldG(z) \right]
  \bolds^0(z) + \frac{1}{2} \left[ \boldH(z)^* \boldH(-z) +
    \boldG(z)^* \boldG(-z) \right] \bolds^0(-z).
\end{equation*}

If we want perfect reconstruction (that is, $\hat\bolds^0 =
\bolds^0$), we need
\begin{equation}\label{eq:pr}
  \begin{gathered}
    \boldH(z)^* \boldH(z)  + \boldG(z)^* \boldG(z)  = 2I, \\
    \boldH(z)^* \boldH(-z) + \boldG(z)^* \boldG(-z) = 0.
  \end{gathered}
\end{equation}

The orthogonality relations \eqref{eq:02}, expressed in terms of
symbols, are
\begin{equation}\label{eq:ortho3}
   \begin{gathered}
     \boldH(z) \boldH(z)^* + \boldH(-z) \boldH(-z)^* = 2I, \\
     \boldG(z) \boldG(z)^* + \boldG(-z) \boldG(-z)^* = 2I, \\
     \boldH(z) \boldG(z)^* + \boldH(-z) \boldG(-z)^* = 0
   \end{gathered}
\end{equation}

It is easy to show that eqs.~\eqref{eq:pr} and \eqref{eq:ortho3} are
equivalent.

\subsection{Multiwavelet Denoising}\label{subsec:denoising}

One of the most interesting signal processing applications is {\em
  denoising}.  In most real applications the original signals are
corrupted by noise. In addition, truncation error in the computation
and in other processing introduces more errors. These errors can be
interpreted as {\em additive white Gaussian noise} (AWGN) $e \sim
N(0,\sigma^2)$ added to the original signal $s$. A {\em denoising
  technique} is a method of removing the noise while retaining
important features of the signal. We consider denoising methods based
on the  multiwavelet transform.

The wavelet transform concentrates the energy of the true signal in a
few wavelet coefficients, while the noise energy will be scattered
among all coefficients. The wavelet coefficients of Gaussian noise are
still Gaussian. Shrinking the small coefficients towards zero will
eliminate most of the noise, while not affecting the signal very much.

We use the following notation. The original signal is $s$. This could
be a one-dimensional signal $s(t)$, or an image $s(x,y)$. The noisy
signal is $\tilde s = s + e$. The denoised signal is $\hat s$. We
likewise use tildes and hats to denote the wavelet coefficients of
$\tilde s$, $\hat s$.

\subsubsection{Stochastic Sequences and Covariance Matrix}

Assume that the input is a noisy signal $\tilde s$. This is sampled
and preprocessed into a vector sequence $\boldstilde^0$. The
preprocessing is described by a linear operator $\calM$. We also
assume that postprocessing $\calN$ is the inverse of preprocessing:
$\calN\calM = I$. The discrete multiwavelet transform (DMWT) is
described by a linear map $\boldTheta$.  For orthogonal multiwavelet
filter banks we have $\boldTheta^T \boldTheta = \boldTheta
\boldTheta^T = I$.

It is important note the influence of the type of transform
(orthonormal or biorthogonal) on the variance of stochastic sequences
subjected to preprocessing and the DMWT. If the preprocessing $\calM$
and $\boldTheta$ are orthonormal transforms, then the energy (i.e.,
the sum of squares of the elements) in the sequence $\calM
\boldstilde$ is unchanged: $\|\boldTheta \calM \boldstilde\|^2 =
\|\calM \boldstilde\|^2 = \|\boldstilde\|^2$. For biorthogonal
transforms, both preprocessing and matrix transformation will change
the input energy.

The noisy signal $\tilde s$ is the sum $\tilde s = s + e$, where $s$
is the true (deterministic) signal (noise free), and $e \sim
N(0,\sigma^2)$ is the noise, with noise power $\sigma^2$. The goal of
signal denoising is to minimize the mean square error $\mbox{MSE} =
\frac{1}{N^2} \|s - \hat s\|$ subject to the condition that the
denoised signal $\hat s$ is at least as smooth as $s$. Due to better
signal representation by the multiwavelet transform, noise and signal
can be separated much more easily. Multiwavelet denoising using a
multivariate shrinkage operator effectively exploits the statistical
information of the transformed wavelet coefficient vectors of noise,
which improves the denoising performance. As a result of the
decomposition and downsampling, the dependence of coefficients is
reduced with increasing levels. The multiwavelet transform
$\boldTheta$ with prefilter $\calM$ of the noisy signal $\tilde s$
\begin{equation*}
  \boldTheta \calM \tilde s = \boldTheta \calM s + \boldTheta \calM e
\end{equation*}
produces the vector wavelet coefficients
\begin{equation*}
  \bolddtilde_k^j = \boldd_k^j + \boldf_k^j,
\end{equation*}
which are correlated with their neighbors.

The scaling function coefficients $\bolds_k^J$ of the signal are
fairly large, and do not need to be thresholded. We can concentrate on
the wavelet coefficients $\boldd_k^j$ of the signal, and $\boldf_k^j$
of the noise.

The sequence $\boldf^j = \{\boldf_k^j\}$ of wavelet coefficients at
level $j$ of the stochastic noise has a multivariate normal
(Gaussian) distribution $\boldf^j \sim N(0,Y_j)$ with a covariance
matrix
\begin{equation*}
  \cov(\boldf^j) = \boldTheta_j \calM_j \Sigma \calM_j^T \boldTheta_j^T,
\end{equation*}
where $\Sigma$ is the covariance matrix of the noise. The matrices
$Y_j$ are $r \times r$ blocks indicating the correlation
between the vector coefficients inside each single decomposition level
\cite{BP03}. In the case of AWGN noise $e \sim N(0,\sigma_e^2)$,
\begin{equation*}
  \cov(\boldf^j) = \boldTheta_j \left[ \sigma_e^2 \calM \calM^T \right]
  \boldTheta_j^T = \sigma_e^2 \boldTheta_j \left[ \calM \calM^T \right] \boldTheta_j^T,
\end{equation*}
the covariance of the stochastic part depends only on the
preprocessing $\calM$ and the choice of multiwavelet transform
$\boldTheta$. If the preprocessing matrix satisfies $\calM \calM^T =
I$, the covariance matrix is $\sigma_e^2 I$. In the general
case, $\cov(\boldf^j) \ne I$ is a symmetric block Toeplitz matrix.

Assuming that the blocks $Y_j$ describing the correlation between the
vector coefficients are known, the nonnegative scalar values
\begin{equation*}
  w_k^j = \sqrt{(\bolddtilde_k^j)^T Y_j^{-1} \bolddtilde_k^j}
\end{equation*}
can be used for thresholding, as proposed \cite{BP03}. In \cite{DS98}
it is shown that it is these values that should be thresholded, and
the wavelet coefficient vectors can then be adapted accordingly.

\subsubsection{Thresholding Rules}

There exist threshold and non-threshold wavelet shrinkage methods. A
non-threshold wavelet shrinkage method is Besov shrinkage
\cite{DGG10,DGG07}.  It is known that its scale is closed under real
interpolation, and it is highly adaptive to the local smoothness.

In this paper, we consider threshold methods in more detail.  This is
a general wavelet denoising procedure \cite{RV91,SAA08}.  Wavelet
shrinkage denoising has been proven to be nearly optimal in the mean
square error (MSE) sense, and performs well in simulation studies.

Thresholding methods can be grouped into two categories \cite{SR04}:
global thresholds and level-dependent thresholds. The first means that
we choose a single value $\lambda$ for the threshold, to be applied
globally to all wavelet coefficients; the second means that a possibly
different threshold value $\lambda_j$ is chosen for each resolution
level $j$. In what follows, we consider a global threshold $\lambda$.

The threshold distinguishes between insignificant coefficients
attributed to noise, and significant coefficients that encode
important signal features. Signals with sparse or near sparse
representation, where only a small subset of the coefficients
represent all or most of the signal energy, are fit for
thresholding. The noise is estimated by a properly chosen threshold
after the multiwavelet transform.

Filtering of additive Gaussian noise (zero mean and standard deviation
$\sigma$) via the thresholding of wavelet coefficients was pioneered
by Donoho \cite{D95}. A wavelet coefficient $\bolddtilde_k^j$ is
compared with a universal threshold $\lambda$. If the coefficient has
a magnitude smaller than the threshold, it is considered to be noise,
and is set to zero. Otherwise, is kept or modified, depending on the
thresholding rule.

In this paper hard and soft thresholding methods will be considered.
The vector wavelet coefficients thresholded by the soft thresholding
rule are
\begin{equation*}
  \bolddhat_k^j = 
  \begin{cases}
    \bolddtilde_k^j \, \dfrac{w_k^j - \lambda}{w_k^j}, & w_k^j \ge \lambda, \\
      0, & w_k^j < \lambda.
  \end{cases}
\end{equation*}
For the hard thresholding rule, they are 
\begin{equation*}
  \bolddhat_k^j = 
  \begin{cases}
    \bolddtilde_k^j, & w_k^j \ge \lambda, \\
    0, & w_k^j < \lambda.
  \end{cases}
\end{equation*}
Another method, semisoft thresholding, has been proposed
\cite{NS07,ZHZS17}, but so far only for the scalar wavelet setting. 

The vector wavelet coefficients $\boldd_k^j$ could be thresholded
using either the hard or soft thresholding rules.  These thresholding
rules have strong connections with model selection in the traditional
linear models.  The trick of thresholding is to include these
irregularities while excluding the noise \cite{DDJ95}.

In general, in terms of visual quality, the noise reduction effect of
soft thresholding is the best, followed by the improved semisoft
shrinkage method; the hard thresholding method is the worst.

Hard shrinkage produces smaller bias but larger variance than soft
shrinkage, and empirically performs better than soft shrinkage minimax
estimation \cite{BG96}.  It better retains local characteristics of
the image edges \cite{D95,S-98a}, and is suitable for denoising of
noise with sudden changes. However, it introduces sudden jumps in the
multiwavelet domain that in turn lead to oscillations in the
reconstructed images. These oscillations are very close to the Gibbs
phenomenon (see fig. 3 in \cite{DF-01}) and are called {\em
  pseudo-Gibbs phenomena}. They affect the visual quality of the
denoising image, and manifest themselves as annoying artifacts in the
denoised image (see Fig. \ref{fig:02}(a)).

The soft thresholding method smoothes the jumps in the wavelet domain
by shrinking also the coefficients near the jumps. The denoised
signals end up relatively smooth, and the pseudo-Gibbs phenomenon is
reduced to some extent.

Better results are expected in the future by using the multiple
description approach for the multiwavelet transformation shown in
\cite{WGLP19}.

\subsubsection{Universal Threshold}

The threshold value $\lambda$ is a very important parameter, which
directly influences the performance of wavelet denoising. It can be
determined by estimating the variance of the noise in images, but
thise value is often unknown in applications.

Using a universal threshold value, instead of a different threshold at
each level, is attractively simple, but is strictly suitable only when
thresholding AWGN wavelet coefficients of variance $\sigma^2$ and mean
zero. The basic idea is that if the signal component is in fact zero,
then with high probability the combination of (zero) signal plus noise
should not exceed the threshold level $\lambda = \sigma_e \sqrt{2 \log
  N}$, where $N$ is the length of the signal.  The universal threshold
$\lambda$ \cite{D95,DDJ95} pays attention to smoothness rather than to
minimizing the mean square error (MSE).

This threshold is the asymptotic infimum, the asymptotic optimal
threshold \cite{DJ94}, and comes close to the minimax threshold, i.e. the
threshold that minimizes the worst case MSE \cite{DDJ98}. The trivial
extension of the universal threshold to multiple wavelet coefficients
is to use the threshold above applied to each coefficient element.

In the numerical examples in this paper, we are using the following
Algorithm 1. It implements the algorithm described in the text,
adapted to two-dimensional signals (images).

\vspace{0.1in}

{\small
\begin{tabular}{ll}
  \hline
  \multicolumn{2}{c}{{\bf Algorithm 1:} Image Denoising Algorithm Based on Multiwavelet Transform} \\
  \hline
  {\bf Step 1} &
    \begin{minipage}[t]{3.5in}
      Apply the pre-filtering $\calM$ to all rows, then all columns (of
      previously row pre-filtered pixels) of the noisy image $\tilde s$
    \end{minipage} \\ 
  {\bf Step 2} &
    \begin{minipage}[t]{3.5in}
      Apply the 2D forward multiwavelet transform row by row and
      column by column, up to level $J$
    \end{minipage} \\ 
  {\bf Step 3} & 
    \begin{minipage}[t]{3.5in}
      Choose the universal threshold $\lambda$ and apply a ({\em hard}
      or {\em soft}) thresholding procedure to the resulting
      multiwavelet coefficients $\bolddtilde_k^j$ to obtain the
      denoised wavelet coefficients $\bolddhat_k^j$
    \end{minipage} \\ 
  {\bf Step 4} & 
    \begin{minipage}[t]{3.5in}
      Apply the 2D backward multiwavelet transform up to level $J$
    \end{minipage} \\ 
  {\bf Step 5} & 
    \begin{minipage}[t]{3.5in}
      Apply the post-filtering $\calN$, producing a denoised image
      $\hat s$
    \end{minipage} \\ 
  \hline
\end{tabular}
}

\subsection{The Multiwavelet SA1}
\label{subsec:SA1}

In this section, we describe a simple multiwavelet that will be used
in the remainder of this paper to illustrate our approach.  We are
interested in supercompact multiwavelets, that is, multiwavelets with
support in $[0,1]$. Such functions have many advantages:
\begin{itemize}
  \item Application to functions defined on a finite interval does not
    require any special treatment at the boundaries of the interval
    \cite{BW-00};
  \item	Interpolation with non-equally spaced data is possible
    \cite{DGSS-99,H-75};
  \item	Application to functions that are only piecewise continuous
    (internal boundaries) can be efficiently implemented
    \cite{BW-00,MT-14,PV-99};
  \item It is easy to construct a nodal basis with (spectral) finite
    elements \cite{BS-07,SM-99};
  \item	Polynomials are reconstructed exactly \cite{BW-00};
  \item Discontinuities in the signal are easy to detect
    \cite{W-1910}, and the size of jumps at element boundaries can be
    measured, in 1D \cite{V-14} and 2D \cite{VR-16};
  \item Construction of the operation matrices in numerical
    methods is simple \cite{AVMN-09,DV-17};
  \item Supercompact multiwavelets with the vanishing moment property
    are suitable for adaptive solvers of PDEs subject to boundary
    conditions, as well as for solving of a wide class of
    integro-differential operators having sparse representation in
    these bases.
\end{itemize}

\begin{figure}
  \includegraphics[width=4.8in]{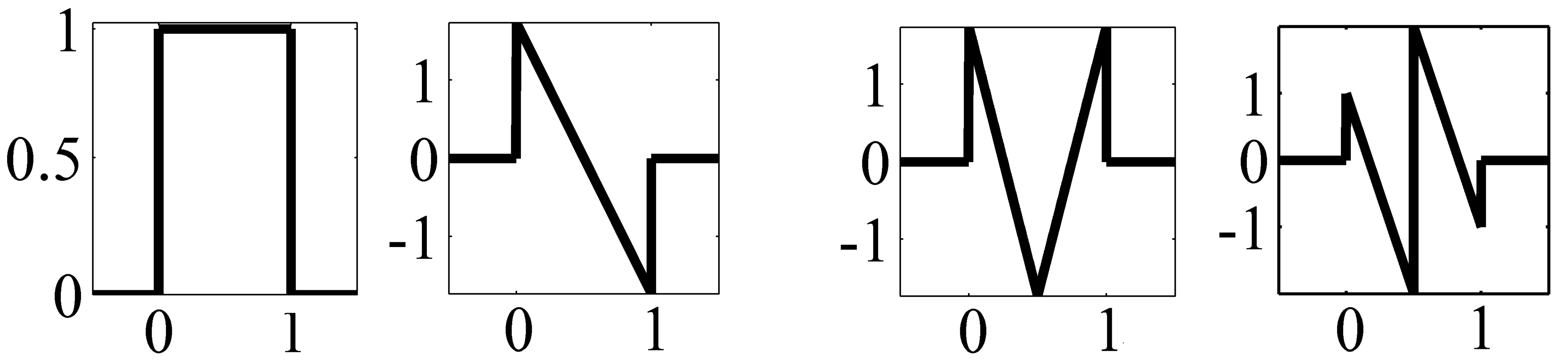} \\
  (a) \hspace{2.2in} (b)
  \caption{The scaling and wavelet functions of SA1;
    (a) the multiscaling function $\boldphi(t) = [\varphi_0(t), \varphi_1(t)]^T$;  
    (b) the multiwavelet function $\boldpsi(t) = [\psi_0(t), \psi_1(t)]^T$.}
  \label{fig:02}
\end{figure}

Alpert \cite{A-90} already observed that a multiscaling function whose
entries form a basis of polynomials of degree $p$ on $[0,1]$ will
always be refinable. For $p=0$, this leads to the Haar wavelet. For
$p=1$, we can take $\phi_0$ = Haar wavelet, and $\phi_1$ = any
non-constant linear polynomial. If we want $\phi_1$ to be orthogonal
to $\phi_0$ and normalized, we find that for $t \in [0,1]$,
\begin{align*}
  \phi_0(t) &= 1, \\
  \phi_1(t) &= \sqrt{3} \, (1 - 2t).
\end{align*}
This choice is unique, up to sign. The two components of $\boldphi$
are shown in fig.~\ref{fig:02}(a).

The refinement coefficients are easy to derive with elementary
algebra: we find
\begin{equation}\label{eq:H}
  H_0 = \frac{\sqrt{2}}{4}
  \begin{bmatrix}
    2 & 0 \\
    \sqrt{3} & 1
  \end{bmatrix}, \quad H_1 = \frac{\sqrt{2}}{4} 
  \begin{bmatrix}
    2 & 0 \\
    -\sqrt{3} & 1
  \end{bmatrix}.
\end{equation}

The completion with a corresponding multiwavelet function is not
unique, but by imposing orthogonality and symmetry constraints we find
an essentially unique completion with recursion coefficients
\begin{equation}\label{eq:G}
  G_0 =  \frac{\sqrt{2}}{4}
  \begin{bmatrix}
    0 & 2 \\
    -1 & \sqrt{3}
  \end{bmatrix}, \qquad
  G_1 = \frac{\sqrt{2}}{4}
  \begin{bmatrix}
    0 & -2 \\
    1 & \sqrt{3}
  \end{bmatrix}.
\end{equation}
Details are given in section \ref{subsec:wavelet}. The corresponding
functions are described by
\begin{align*}
  \psi_0(t) &=
  \begin{cases}
    \sqrt{3} \, (1-4t) & t \in [0,\frac{1}{2}], \\
    \sqrt{3} \, (4t-3) & t \in [\frac{1}{2},1],
  \end{cases} \\
  \psi_1(t) &=
  \begin{cases}
    (1-6t) & t \in [0,\frac{1}{2}], \\
    (5-6t) & t \in [\frac{1}{2},1].
  \end{cases}
\end{align*}
They are shown in fig.~\ref{fig:02}(b).

We call this multiwavelet SA1.

\section{Matrix Product Filters}
\label{sec:mpfdesign}

\subsection{Theory of Matrix Product Filters}

Given a matrix filter
\begin{equation*}
  \boldH(z) = \sum_{k=0}^m H_k z^{-k},
\end{equation*}
its {\em matrix product filter} $\boldP(z)$ is defined as
\begin{equation*}
  \boldP(z) = \boldH(z) \boldH(z)^* = \sum_{k=-m}^m P_k z^k,
\end{equation*}
where
\begin{equation*}
  P_k = H_0 H_k^T + \cdots + H_{m-k-1} H_{m-1}^T + H_{m-k} H_m^T.
\end{equation*}
In particular for $k = 0$, 
\begin{equation*}
    P_0 = H_0 H_0^T + \cdots + H_{m-1} H_{m-1}^T + H_{m} H_m^T.
\end{equation*}

The Laurent polynomial matrix $\boldP(z) \in \C^{r \times r}[z,
  z^{-1}]$ is a discrete-time para-Hermitian matrix. Here $\C^{r
  \times r}[z, z^{-1}]$ is the ring of $r \times r$ matrices whose
elements are polynomials in $z$, $z^{-1}$ with complex coefficients
\cite{B-17}. By the first equation in \eqref{eq:ortho3}, it is a {\em
  half-band filter}~\cite{CNKS-96}, which means it satisfies the
equation
\begin{equation}\label{eq:03}
  \boldP(z) + \boldP(-z) = 2I.
\end{equation}
This implies
\begin{align*}
  P_0 &= I, \\
  P_{2k} &= 0 \qquad \mbox{for $k \ne 0$}.
\end{align*}

We can similarly define the half-band filter
\begin{equation*}
  \boldQ(z) = \boldG(z) \boldG(z)^*.
\end{equation*}

Every orthogonal multiscaling filter $\boldH(z)$ is a spectral factor
of some half-band multifilter. We use this fact as the starting point
for constructing multiwavelets: Construct a suitable $\boldP(z)$
first, and factor it to obtain $\boldH(z)$.

In the scalar case, spectral factorization is one of the main
design techniques~\cite{BPDS18,SN-96}. For $r>1$, matrix spectral
factorization is required, which is more challenging.

\subsection{Matrix Spectral Factorization}

A critical component in classical spectral decomposition is the {\em
  Fej{\'e}r-Riesz theorem} for positive definite functions
\cite{C-1907}. Fej{\'e}r \cite{F-1916} was the first to note the
importance of the class of trigonometric polynomials that assume only
non-negative real values; the theorem was considered and proved by
Riesz \cite{R-1916}.

The Fej{\'e}r-Riesz theorem in one dimension considers a trigonometric
polynomial expressed in the form
\begin{equation*}
  \nu(z) = \sum_{k=-N}^N \nu_k z^k.
\end{equation*}

When $\nu(z)$ is real for all $z \in \T$, the coefficients satisfy
$\overline{\nu_k} = \nu_{-k}$ for all $k$. The theorem states that if
$\nu(z) \ge 0$ for all $z \in \T$, such a $\nu(z)$ is expressible in
the form
\begin{equation*}
  \nu(z) = p(z) p(z)^*
\end{equation*}
for some polynomial $p(z) = \sum_{k=0}^N p_k z^k$. As noted in
\cite{EJL07}, a spectral factor $p(z)$ is unique up to a constant
right unitary multiplier $U(z)$, i.~e.
\begin{equation*}
  p_{new}(z) = p(z) U(z).
\end{equation*}

The Fej{\'e}r-Riesz theorem does not extend to factorization of
multivariate trigonometric polynomials (see \cite{D-04} for some
counterexamples).  Relevant key theorems are \cite[Theorem 3.2]{R-63}
in the 1D case, \cite[Theorem 6.2]{D-04} for the 2D case, with
examples and necessary and sufficient conditions in \cite{GW-04}, and
\cite[Theorem 3.1]{R-63} for arbitrary dimensions.

However, the theorem can be extended to the case of univariate
polynomials with matrix coefficients. This is {\em Matrix Spectral
  Factorization} (MSF) (see \cite{V-72,C-85,ESS-18,YK-78}).

The MSF problem plays a crucial role in the solution of various
applied problems for MIMO systems in communications and control
engineering \cite{B-17,KE-15,WMW-15,YZC-15}.

\subsection{Bauer's Method}

A well-known method for MSF is Bauer's method \cite{B-56}. This method
has been successfully applied in \cite{C-07,DXD-13,F-05,HCW-10}.
Details of the algorithm are given in section \ref{subsec:msf} below.

Bauer's method, in the implementation of Youla and Kazanjian
\cite{MS18,M13,M07,YK-78}, has the advantage over other approaches
that it can handle zeros of the determinant of $\boldP(z)$ on $\T$.
Unfortunately, the presence of these zeros affects the accuracy and
speed of convergence.  In its original form, the method only gives us
a low precision spectral factor, and convergence is very slow.

The approaches in \cite{ESS-18} for finding precision spectral factors
can be used instead. For instance, the Janashia-Lagvilava MSF method
for the orthogonal SA4 multiwavelet considered in \cite{ESS-18} is
exact.

In this paper, we achieve an exact MSF leading to a simple orthogonal
multiwavelet by the closed form of Bauer’s method for short product
filters. This is possible for some types of short support
multiwavelets, especially the supercompact multiwavelets \cite{BW-00}.

\subsection{A Simple Matrix Product Filter}
\label{subsec:simple}

The simplest non-scalar example is for $r = 2$, $m = 1$. In this case,
$\boldP \in C^{2 \times 2}[z, z^{-1}]$ has the form
\begin{equation}\label{eq:Pdef}
  \boldP(z) = P_1^T z^{-1} + P_0 + P_1 z.
\end{equation}
From the half-band condition \eqref{eq:03} we see that $P_0$ is the
identity matrix.

The smoothness of the multiscaling and multiwavelet functions is
improved if the determinant of the product filter has a higher-order
zero at $z = -1$ (see \cite{V-86}). Therefore, we choose
\begin{equation}\label{eq:05}
  \det(\boldP(z)) = \left( \frac{1 + z^{-1}}{2} \right)^k q(z),
\end{equation}
where $q(z)$ is a linear phase polynomial in $z$, and $k$ is
necessarily even.

The simplest case corresponds to $k=4$ and $q(z) = z^2$, where
\begin{equation*}
  \det(\boldP(z)) = \left( \frac{1 + z^{-1}}{2} \right)^4 z^2 =
  \frac{1}{16} \left( z^{-2} + 4 z^{-1} + 6 + 4 z + z^2 \right).
\end{equation*}
Setting
\begin{equation*}
  P_1 =
  \begin{bmatrix}
    a & b \\
    c & d
  \end{bmatrix},
\end{equation*}
we find the equations
\begin{align*}
  ad - bc &= 1/16, \\
  a+d &= 4/16, \\
  1 - b^2 - c^2 + 2ad &= 6/16.
\end{align*}

These equations have precisely four solutions: either $a = 1/2$, $d =
-1/4$ or $a = -1/4$, $d = 1/2$, and in both cases we can choose $b =
-c = \pm \sqrt{3}/4$. However, these are all essentially equivalent.
The different choices just correspond to interchanging $\phi_0$ and
$\phi_1$, or changing the sign of one of the functions.

We choose
\begin{equation*}
  a = \frac{1}{2}, \qquad b = -\frac{\sqrt{3}}{4}, \qquad c =
  \frac{\sqrt{3}}{4}, \qquad d = -\frac{1}{4}.
\end{equation*}

The resulting product filter has coefficient
\begin{equation}\label{eq:P1}
  P_1 = \frac{1}{4}
  \begin{bmatrix}
    2 & - \sqrt{3} \\
    \sqrt{3} & -1
  \end{bmatrix}.
\end{equation}

\section{Design of Orthogonal Multiwavelet Filter}
\label{sec:owmdesign}

\subsection{Finding $\boldH(z)$}
\label{subsec:msf}

For simplicity, we only describe the details of Bauer's method for the
case $m=1$ considered in this paper.  The equation $\boldP(z) =
\boldH(z) \boldH(z)^*$ is equivalent to the fact that the infinite
block tridiagonal matrix
\begin{equation*}
  P =
  \begin{bmatrix}
    \ddots & \ddots & \ddots \\
    & P_1^T & P_0 & P_1 \\
    & & P_1^T & P_0 & P_1 \\
    & & & \ddots & \ddots & \ddots
  \end{bmatrix}
\end{equation*}
can be factored as $P = H H^T$, where $H$ is the infinite block
bidiagonal matrix
\begin{equation*}
  H =
  \begin{bmatrix}
    \ddots & \ddots \\
    & H_1 & H_0 \\
    & & H_1 & H_0 \\
    & & & \ddots & \ddots
  \end{bmatrix}
\end{equation*}

For a chosen integer $n$, we truncate $P$ to the matrix $P_n$ of size
$(n+1) \times (n+1)$, and do a Cholesky factorization $P_n = H_n
H_n^T$. The last row of $H_n$ will be $H_1^{(n)}$, $H_0^{(n)}$.

The YK theory says that $H_0^{(n)} \to H_0$ as $n \to \infty$, and
likewise for $H_1$. The computation can be streamlined, by defining
\begin{equation*}
  X^{(k)} = H_0^{(k)} \left[ H_0^{(k)} \right]^T, \quad k = 0, 1, \ldots.
\end{equation*}
The Cholesky factorization algorithm is equivalent to 
\begin{equation}\label{eq:fpi}
   \begin{split}
     X^{(0)} &= P_0, \\
     X^{(k+1)} &= P_0 - P_1^T \left[ X^{(k)} \right]^{-1} P_1.  
   \end{split}
\end{equation}
The limit matrix $X$ satisfies
\begin{equation}\label{eq:X}
  X = P_0 - P_1^T X^{-1} P_1.
\end{equation}
We can use fixed point iteration based on \eqref{eq:fpi} to avoid
forming the large matrix $P_n$. Once $X$ has been found to sufficient
accuracy, we can find $H_0$ as the Cholesky factor of $X$, and $H_1 =
P_1^T H_0^{-T}$.

Even better, in the simple case considered in this paper, the Symbolic
Math toolbox in Matlab is able to solve equation \eqref{eq:X} in
closed form. On a MacBook Air with 2.2GHz Intel i7 processor, using
Matlab 9.1, the computational time is under 0.8 sec. We get the exact
matrix spectral factor with coefficients
\begin{equation}\label{eq:H2}
  H_0 = \frac{\sqrt{2}}{4}
  \begin{bmatrix}
    2 & 0 \\
    \sqrt{3} & 1
  \end{bmatrix}, \quad H_1 = \frac{\sqrt{2}}{4} 
  \begin{bmatrix}
    2 & 0 \\
    -\sqrt{3} & 1
  \end{bmatrix}.
\end{equation}

This is precisely the multiscaling function of SA1, introduced in
section~\ref{subsec:SA1}.

The spectral factor obtained by our closed form of Bauers method is
precise. This multiscaling function can also be obtained from the 4x4
Type-II DCT matrix \cite{GM-05} and by using the Janashia-Lagvilava
method \cite{ESS-18} in exact form. It can also be obtained
approximately by the Janashia-Lagvilava matrix spectral factorization
method (see procedure 2 in \cite{JLE-11}), with an error in the
spectral factors of about $10^{-8}$.

\subsection{Finding $\boldG(z)$}
\label{subsec:wavelet}

The orthogonality conditions~\eqref{eq:02} in the case $m=1$ simply
state that the matrix
\begin{equation*}
  \begin{bmatrix}
    H_0 & H_1 \\
    G_0 & G_1
  \end{bmatrix}
\end{equation*}
must be orthogonal. We can find an initial orthogonal completion
$[\hat G_0, \hat G_1]$ from a QR-decomposition of $[H_0,
  H_1]^T$. Matlab returns
\begin{equation*}
  [\hat G_0, \hat G_1] = \frac{\sqrt{2}}{4}
  \begin{bmatrix}
    1 & -\sqrt{3} & -1 & -\sqrt{3} \\
    0 & 2 & 0 & -2
  \end{bmatrix}.
\end{equation*}
Any other orthogonal completion has to be of the form
\begin{equation*}
  [ G_0, G_1 ] = U \cdot [\hat G_0, \hat G_1]
\end{equation*}
for orthogonal $U$. We want to choose a $U$ that introduces symmetry.

A function $f(t)$ is {\em symmetric} about a point $a$ if $f(a+t) =
f(a-t)$. It is {\em antisymmetric} about $a$ if $f(a+t) = -
f(a-t)$. For multiscaling and multiwavelet functions, we only consider
the simplest case, where each component $\varphi_i$ or $\psi_i$ is
symmetric or antisymmetric about the midpoint $m/2$. Thus,
\begin{equation}\label{eq:sym1}
  \boldphi(\frac{m}{2} + t) = S \cdot \boldphi(\frac{m}{2} - t),
\end{equation}
where $S$ is a diagonal matrix with entries $1$ (symmetry) or $(-1)$
(antisymmetry). Likewise,
\begin{equation*}
  \boldpsi(\frac{m}{2} + t) = T \cdot \boldpsi(\frac{m}{2} - t).
\end{equation*}

It is not hard to show that $\boldphi$, $\boldpsi$ have the symmetry
given by $S$, $T$ if and only if
\begin{equation}\label{eq:sym2}
   \begin{split}
     H_k &= S H_{m-k} S, \\
     G_k &= T G_{m-k} S, \qquad k=0,\ldots,m.
   \end{split}
\end{equation}

In our example, we observe that the original factors \eqref{eq:H2}
already satisfy
\begin{equation*}
  H_1 = S H_0 S
\end{equation*}
for $S = \diag(1,-1)$. That is, component $\varphi_0$ is symmetric,
$\varphi_1$ is antisymmetric.

The symmetry conditions \eqref{eq:sym2} for $\boldG$ are
\begin{align*}
  U \hat G_0 &= T U \hat G_1 S, \\
  U \hat G_1 &= T U \hat G_0 S,
\end{align*}
which leads to
\begin{equation*}
  U = T U \hat G_1 S \hat G_0^{-1} = - T U S.
\end{equation*}
It is easy to check that $T = \pm I$ is not possible, so $T = \pm
S$. We choose $T = S$, which means that we want $\psi_0$ to be
symmetric, $\psi_1$ antisymmetric.

Any $U$ of the form
\begin{equation*}
  U =
  \begin{bmatrix}
    0 & \pm 1 \\
    \pm 1 & 0
  \end{bmatrix}
\end{equation*}
is then suitable. The choice of signs corresponds to changing the sign of
$\psi_0$ and/or $\psi_1$. Using
\begin{equation*}
  U =
  \begin{bmatrix}
    0 & 1 \\
    -1 & 0
  \end{bmatrix}
\end{equation*}
leads to the same $G_0$, $G_1$ as in \eqref{eq:G}.

Choosing $T = -S$ would interchange $\psi_0$ and $\psi_1$, and give us
\begin{equation*}
  U =
  \begin{bmatrix}
    \pm 1 & 0 \\
    0 & \pm 1
  \end{bmatrix},
\end{equation*}
which again corresponds to sign changes.

\section{Efficient Implementation}
\label{sec:implementation}

\begin{figure}
  (a)
  \includegraphics[height=1.1in]{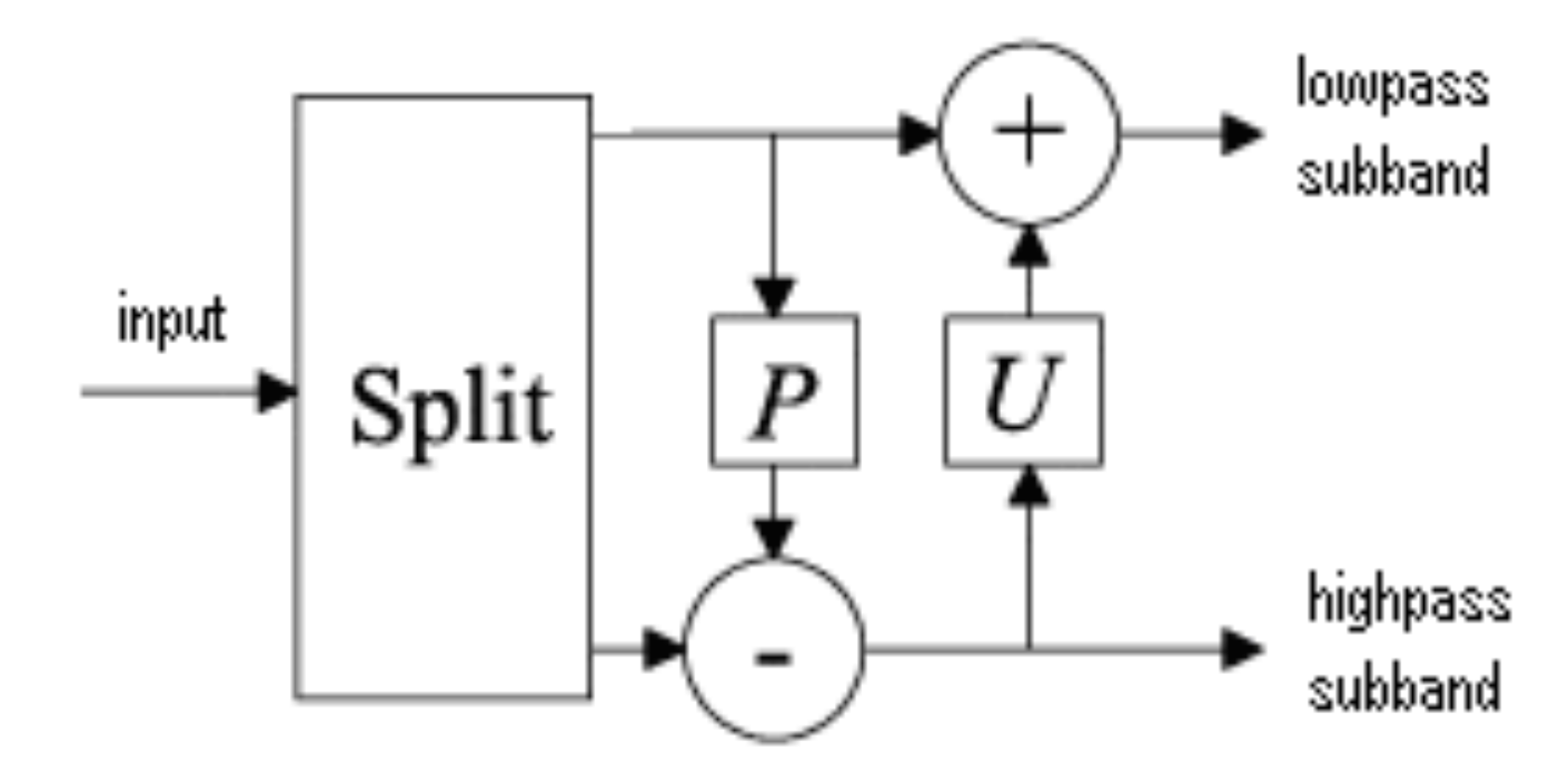} \quad
  (b)
  \includegraphics[height=1.1in]{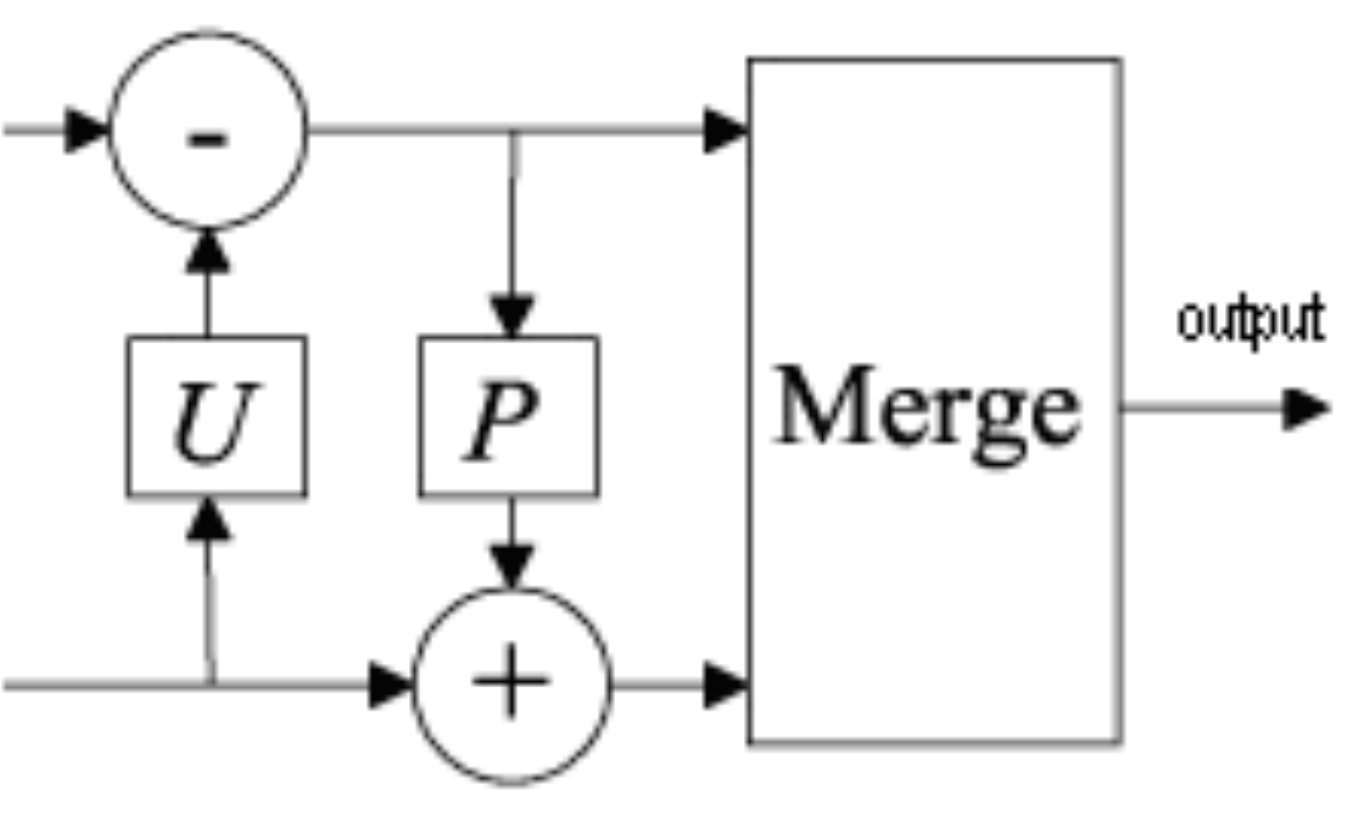}
  \caption{One lifting step; (a) analysis part; (b) synthesis part.}
  \label{fig:03}
\end{figure}

We give here a brief overview of the lifting scheme (LS) and some
rounding operations.  Lifting steps are an efficient implementation of
filtering operations (see fig.~\ref{fig:03}(a)).  One lifting step for
a filter bank consists of an operation {\em Split}, a prediction $P$, and
an update $U$.

One of the great advantages of LS is that it decomposes a multiwavelet
filter into simple invertible steps~\cite{CDSY-98,S-98b}. The inverse
LS can be obtained from the forward LS by reversing the order of the
operations, inverting the signs in the lifting steps, and replacing
the {\em Split} by a {\em Merge} step (see fig.~\ref{fig:03}(b)).

If a {\em floor} (or {\em round} or {\em ceiling}) operator is placed
in each lifting step, the filter banks can now map integers to
integers with perfect reconstruction, regardless of whether the
lifting step has integer coefficients or not.

Moreover, if the factors are chosen to be dyadic, multiplication can
be implemented by addition and binary bit shift. If the integer
mantissa $k$ in a lifting step with coefficient $c = k \cdot 2^{-b_0}$
is written in binary notation as
\begin{equation*}
  k = \sum k_i 2^i, \qquad \mbox{$k_i = 0$ or $1$},
\end{equation*}
then the multiplication of a number $x$ by $c$ can be implemented by
adding copies of $x$, left-shifted by $i$, for each $k_i \ne 0$, and a
final right-shift by $b_0$ bits.

This approach leads to an implementation with solely binary shifts,
preventing bit-depth expansion in intermediate results. With all
scaling factors set to unity, multiplierless filter banks can be
easily constructed using this method. If the scaling factors are
powers of two, we can still retain the multiplierless feature on
filter banks.

The lifting scheme can be done in place: we never need samples other
than the output of the previous lifting step, and therefore we can
replace the old stream by the new stream at every summation point.

\subsection{The Lifting Scheme}
\label{subsec:lifting}

Based on matrix factorization theory~\cite{HS-01}, a nonsingular
matrix can be factored into a product of at most three triangular
elementary PLUS reversible matrices~\cite{SH-05,WWJS-09,WWJS-10}. If
the diagonal elements of the matrices are ones, a reversible
integer-to-integer transform can be realized by LS.

In order to speed up the implementation, it is necessary to optimize
the elementary triangular factorization by either minimizing the
number of factorized matrices or the computational complexity of each
step. The factorization is not unique, and there are other different
factorizations that affect the error between the dyadic approximation
transform and the original transform.

For input vector $\boldx = [x_0, x_1, x_2, x_3]^T$ and output vector
$\boldy = [y_0, y_1, y_2, y_3]^T$, the lifting implementation of the
forward DMWT~\eqref{eq:mwt4} is
\begin{equation}\label{eq:lifting}
  \boldy = H \boldx = (PLU) \boldx = 
  \begin{bmatrix}
    1 & 0 & 0 & 0 \\
    0 & 0 & 0 & 1 \\
    0 & 1 & 0 & 0 \\
    0 & 0 & 1 & 0
  \end{bmatrix}
  \begin{bmatrix}
    1 & 0 & 0 & 0 \\
    0 & 1 & 0 & 0 \\
    -\frac{1}{2} & \frac{\sqrt{3}}{2} & 1 & 0 \\
    \frac{\sqrt{3}}{2} & \frac{1}{2} & -\sqrt{3} & 1
  \end{bmatrix}
  \begin{bmatrix}
    1 & 0 & 1 & 0 \\
    0 & 1 & 0 & -1 \\
    0 & 0 & 1 & \sqrt{3} \\
    0 & 0 & 0 & 4
  \end{bmatrix}
  \begin{bmatrix}
    x_0 \\ x_1 \\ x_2 \\ x_3
  \end{bmatrix}.
\end{equation}

This leads to the synthesis implementation
\begin{equation}\label{eq:invlifting}
  H^{-1} = \frac{1}{2} U^{-1} L^{-1} P^T.
\end{equation}
We include the factor of $(1/2)$ with the reconstruction instead of
the decomposition (see~\eqref{eq:mwt4}).

In our implementation, we write the $L$ and $U$ matrices as products
of elementary triangular and/or diagonal reversible matrices
(see~\cite{Str-06}, page 34), for faster evaluation.
\begin{align*}
  L &=
  \begin{bmatrix}
    1 & 0 & 0 & 0 \\
    0 & 1 & 0 & 0 \\
    -1/2 & \sqrt{3}/2 & 1 & 0 \\
    0 & 0 & 0 & 1
  \end{bmatrix}
  \begin{bmatrix}
    1 & 0 & 0 & 0 \\
    0 & 1 & 0 & 0 \\
    0 & 0 & 1 & 0 \\
    \sqrt{3}/2 & 1/2 & -\sqrt{3} & 1
  \end{bmatrix}, \\
  U &=
  \begin{bmatrix}
    1 & 0 & 0 & 0 \\
    0 & 1 & 0 & 0 \\
    0 & 0 & 1 & 0 \\
    0 & 0 & 0 & 4
  \end{bmatrix}
  \begin{bmatrix}
    1 & 0 & 0 & 0 \\
    0 & 1 & 0 & -1 \\
    0 & 0 & 1 & \sqrt{3} \\
    0 & 0 & 0 & 1
  \end{bmatrix}
  \begin{bmatrix}
    1 & 0 & 1 & 0 \\
    0 & 1 & 0 & 0 \\
    0 & 0 & 1 & 0 \\
    0 & 0 & 0 & 1
  \end{bmatrix}
\end{align*}

\begin{figure}
  (a)
  \includegraphics[height=1.7in]{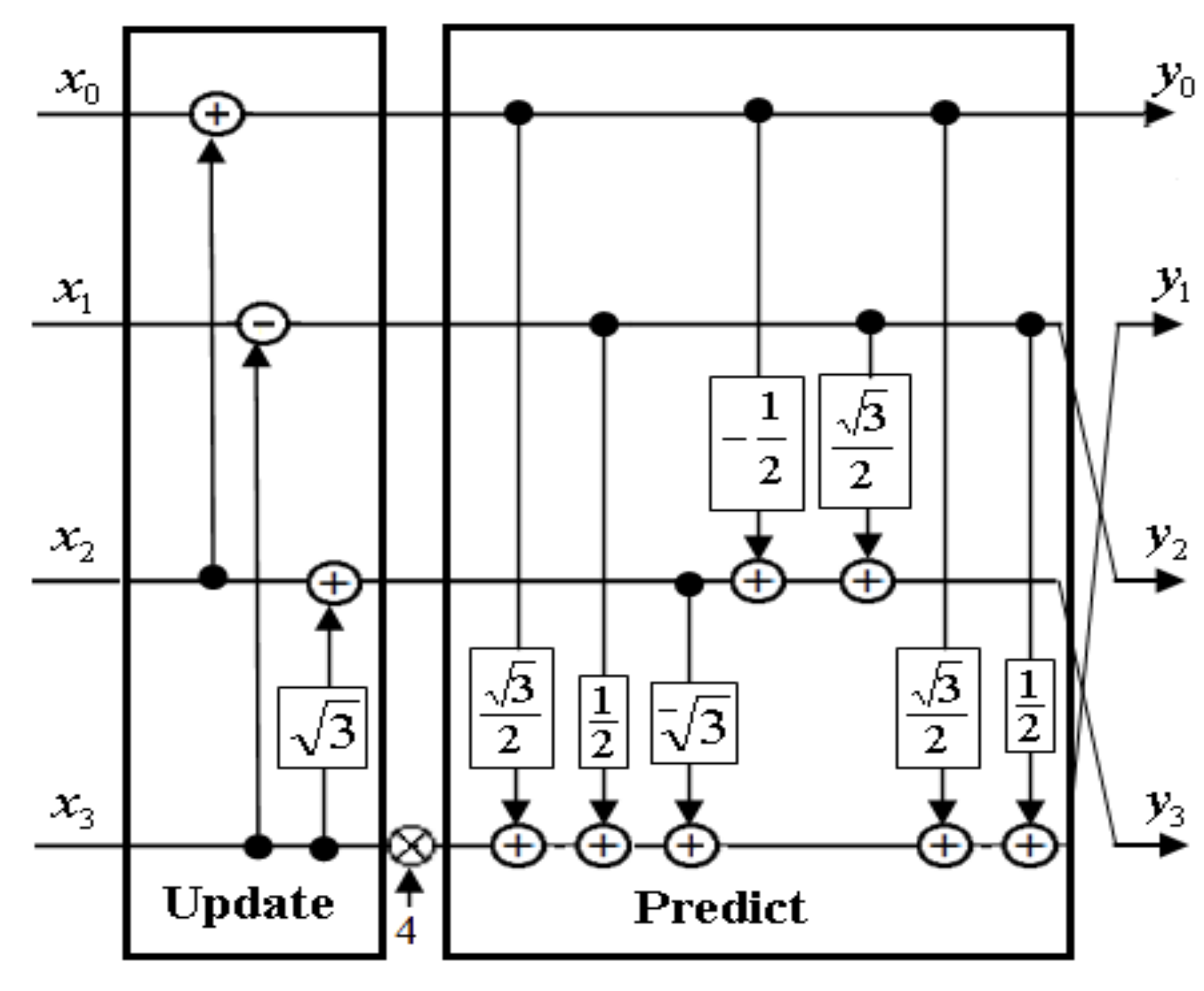} \quad
  (b)
  \includegraphics[height=1.7in]{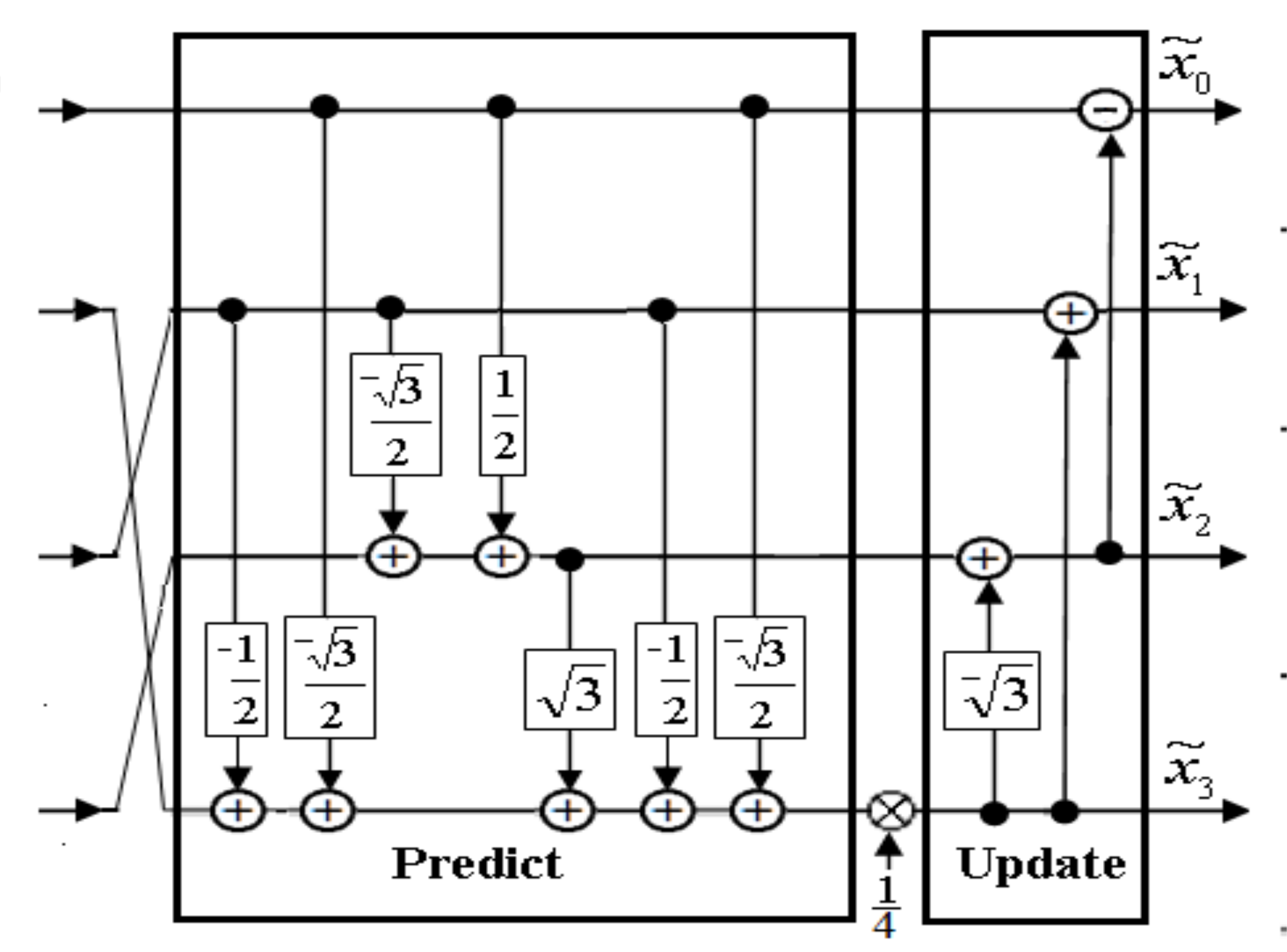}
  \caption{The PLUS matrices implementation of $H$; (a) forward; (b) backward.}
  \label{fig:04}
\end{figure}

The implementation of the lifting scheme is shown in
fig.~\ref{fig:04}.

\subsection{Multiplierless architecture of the coefficient $\sqrt{3}$}\label{subsec:dyadic}

For the transform to be reversible, it must be able to exactly
reconstruct the input from the output. Low computational complexity is
another desirable property.  In order to avoid floating-point
multiplications, we approximate values of the lifting matrices by
dyadic numbers.

The lifting scheme described in subsection~\ref{subsec:lifting} uses
only a single non-dyadic number, which is $\sqrt{3}$. We want to
approximate this coefficient by a number of the form $k \cdot
2^{-b_0}$ with integer $k$. Since $\sqrt{3}$ is between 1 and 2, this
corresponds to an approximation with $(b_0+1)$ bits.  For example, for
$b_0=5$, $\sqrt{3}$ is approximated by the 6-bit number $55 \cdot
2^{-5} = 1.71875$, with an error of approximately $1.73205 - 1.71875 =
0.0133$.

The dyadic approximations of the coefficient $\sqrt{3}$ and its
quantization errors are shown in table \ref{tab:01}.  The
table also shows the number of adders required to implement the
multiplierless structure.

\begin{figure}
  (a)
  \includegraphics[height=1.8in]{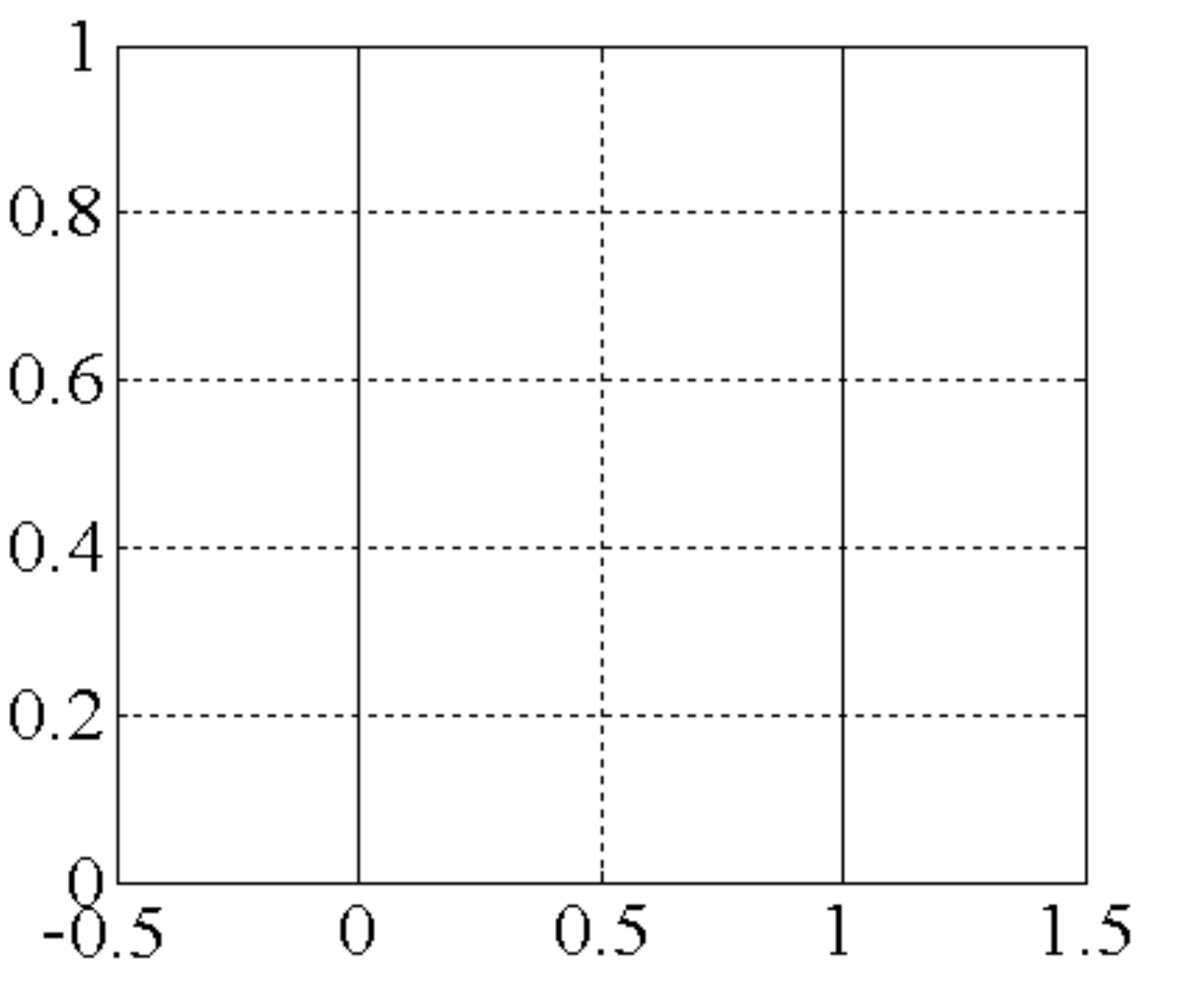} \quad
  (b)
  \includegraphics[height=1.8in]{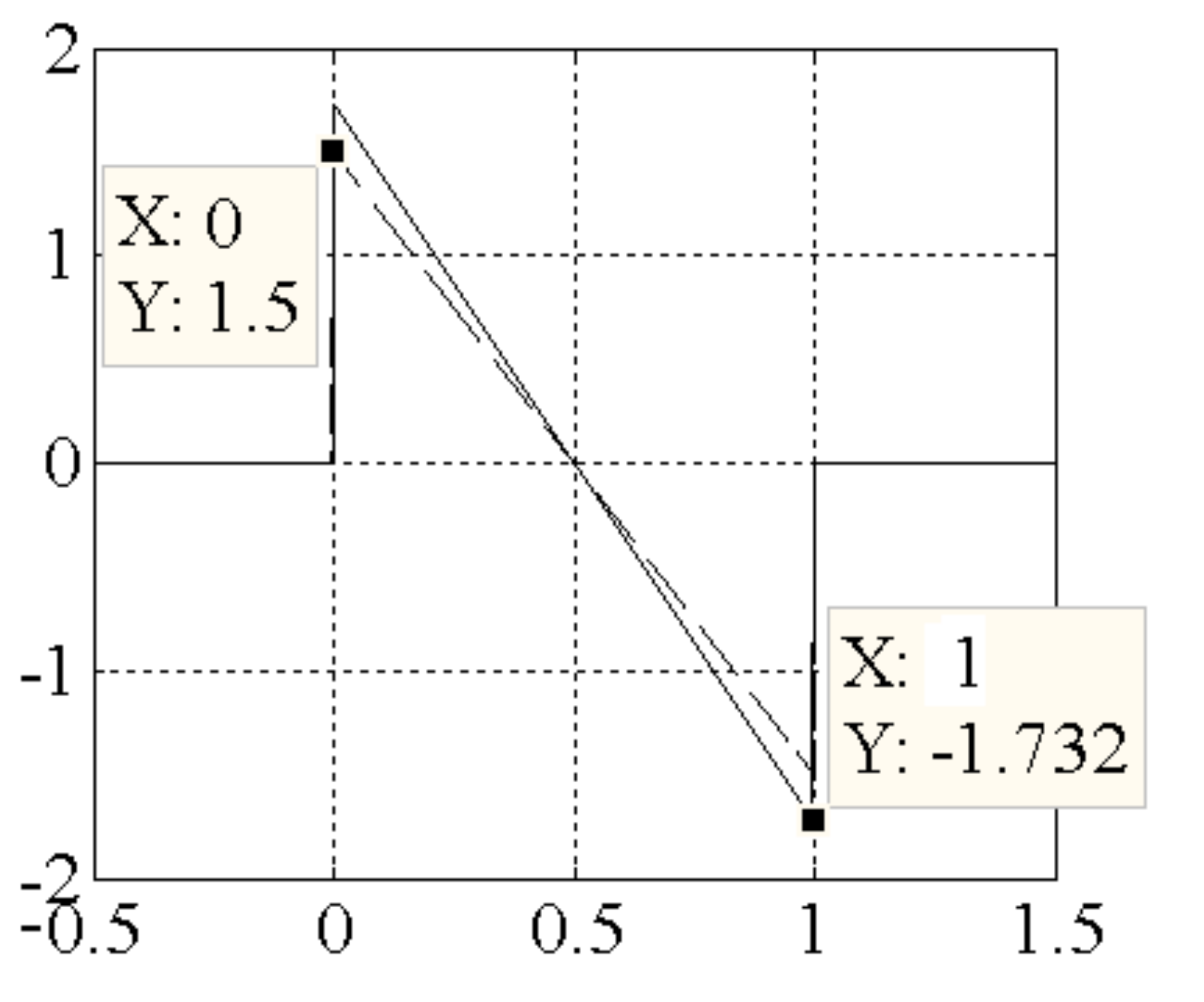} \\
  (c)
  \includegraphics[height=1.8in]{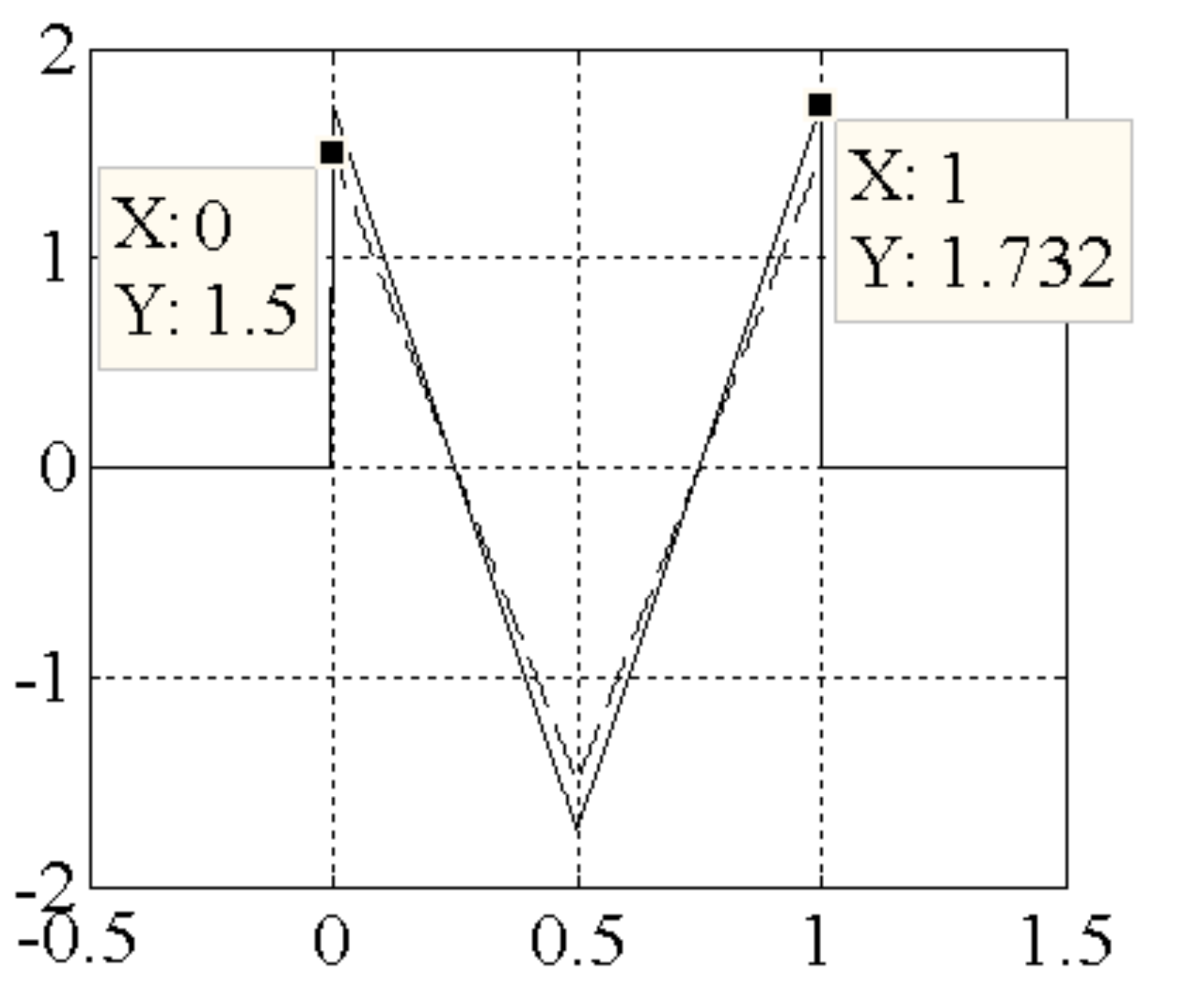} \quad
  (d)
  \includegraphics[height=1.8in]{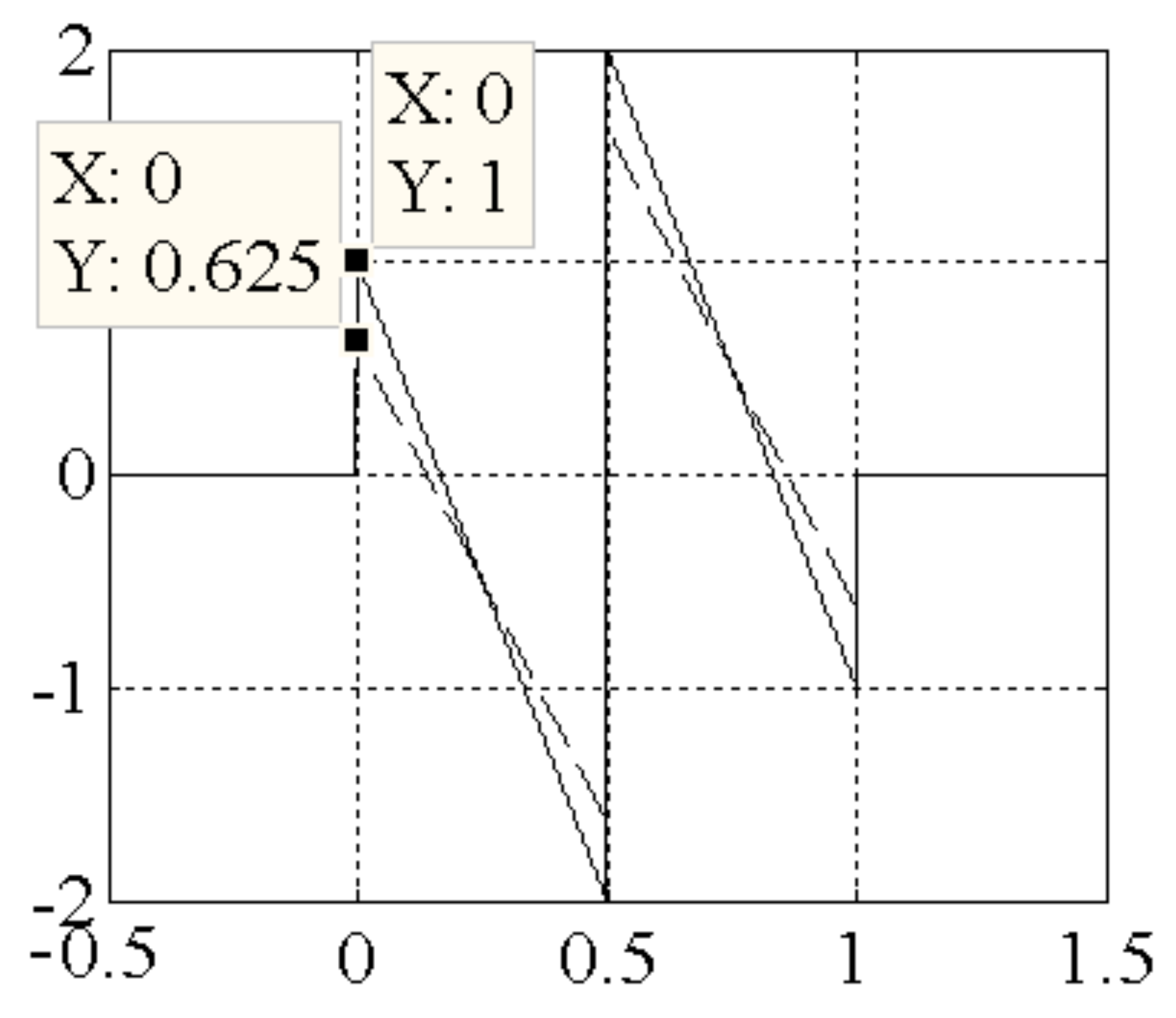}
  \caption{Floating-point arithmetic (solid line) in comparison with
    fixed point arithmetic (dash-dot line) of the impulse responses of
    the SA1 orthogonal multiwavelet, with 2-bit dyadic approximation
    of the coefficient $\sqrt{3}$ (that is, $\sqrt{3} \approx 3/2$);
    (a) $\phi_0$; (b) $\phi_1$; (c) $\psi_0$; (d) $\psi_1$.}
  \label{fig:05}
\end{figure}

We illustrate the effect of quantization for $b_0=1$ in more detail.
This is the case of 2-bit approximation, with $\sqrt{3} \approx 3/2$.

The true recursion coefficients of SA1 are given in eqs. \eqref{eq:H}
and \eqref{eq:G}. The actual recursion coefficients used in the 2-bit
approximation are
\begin{equation}\label{eq:24}
   \begin{gathered}
  \tilde H_0 = \frac{\sqrt{2}}{4}
  \begin{bmatrix}
    2 & 0 \\
    1.5 & 1
  \end{bmatrix}, \quad \tilde H_1 = \frac{\sqrt{2}}{4} 
  \begin{bmatrix}
    2 & 0 \\
    -1.5 & 1
  \end{bmatrix}, \\
  \tilde G_0 = \frac{\sqrt{2}}{4}
  \begin{bmatrix}
    0 & 2 \\
    -1 & 1.5
  \end{bmatrix}, \qquad
  \tilde G_1 = \frac{\sqrt{2}}{4}
  \begin{bmatrix}
    0 & -2 \\
    1 & 1.5
  \end{bmatrix}.
   \end{gathered}
\end{equation}

These approximate matrix coefficients \eqref{eq:24} satisfy
\begin{equation*}
   \begin{gathered}
     \tilde\boldH(z) \tilde\boldH(z)^* + \tilde\boldH(-z) \tilde\boldH(-z)^* =
     \begin{bmatrix}
       2 & 0 \\
       0 & \frac{13}{8}
     \end{bmatrix}, \\
     \tilde\boldG(z) \tilde\boldG(z)^* + \tilde\boldG(-z) \tilde\boldG(-z)^* = 
     \begin{bmatrix}
       2 & 0 \\
       0 & \frac{13}{8}
     \end{bmatrix}, \\
     \tilde\boldH(z) \tilde\boldG(z)^* + \tilde\boldH(-z) \tilde\boldG(-z)^* = 0
   \end{gathered}
\end{equation*}
instead of the true orthogonality conditions \eqref{eq:ortho3}.

This leads to a decrease of CG to $2.01$ dB and non-perfect
reconstruction, which is undesirable in many mathematical and
engineering applications.

The effect of rounding can also be illustrated by its effect on the
multiscaling and multiwavelet functions. It is easy to verify that the
functions which satisfy the rounded recursion relations corresponding
to~\eqref{eq:01} are
\begin{equation*}
  \tilde \phi_0(t) = 1, \qquad 
  \tilde \phi_1(t) = \frac{3}{2} (1-2t)
\end{equation*}
and
\begin{equation*}
  \tilde\psi_0(t) =
  \begin{cases}
    \frac{3}{2} (1-4t) & t \in [0,\frac{1}{2}], \\
    \frac{3}{2} (4t-3) & t \in [\frac{1}{2},1],
  \end{cases}
  \qquad
  \tilde\psi_1(t) =
  \begin{cases}
    \frac{5}{8} - \frac{9}{2} t & t \in [0,\frac{1}{2}], \\
    \frac{31}{8} - \frac{9}{2} t & t \in [\frac{1}{2},1].
  \end{cases}
\end{equation*}
This is illustrated in fig.~\ref{fig:05}.  The biggest differences are
at the end points, which are marked by squares in the figure.

Obviously, the dyadic approximation strongly influences the sloped
parts in time domain, which leads to smaller magnitudes for the
scaling function $\tilde\phi_1$, as well as both wavelet functions
$\tilde\psi_0$ and $\tilde\psi_1$.

\begin{table}
  \caption{Different quantizations of the non-trivial coefficient
    $\sqrt{3} \approx k \cdot 2^{-b_0}$, with $k$ an integer
    represented by $(b_0+1)$ bits. For example, for $b_0=5$, the
    approximation is $\sqrt{3} \approx 55 \cdot 2^{-5} = 1.71875$,
    with an error of approximately $1.73205 - 1.71875 = 0.0133$.  The
    table also shows the number of adders needed to realize the
    multiplierless structure. Entries in bold mark the places where
    increasing the number of bits does not change the approximation.}
  \label{tab:01}
  \begin{tabular}{cccr}
    \hline
    $b_0$ = binary exponent & mantissa & adders &
    \multicolumn{1}{c}{quantization error} \\
    \hline
    1 & 3 & 1 & $0.232050807568877$ \\
    2 & 7 & 2 & $-0.017949192431123$ \\
    {\bf 3} & {\bf 14} & {\bf 2} & $----$ \\
    {\bf 4} & {\bf 28} & {\bf 2} & $----$ \\
    5 & 55 & 2 & $0.013300807568877$ \\
    6 & 111 & 2 & $-0.002324192431123$ \\
    {\bf 7} & {\bf 222} & {\bf 2} & $----$ \\
    8 & 443 & 3 & $ 0.001582057568877$ \\
    9 & 887 & 3 & $-0.000371067431123$ \\
    {\bf 10} & {\bf 1774} & {\bf 3} & $----$ \\
    11 & 3547 & 4 & $0.000117213818877$ \\
    {\bf 12} & {\bf 7094} & {\bf 4} & $----$ \\
    13 & 14189 & 4 & $-0.000004856493623$ \\
    {\bf 14} & {\bf 28378} & {\bf 4} & $----$ \\
    \hline
  \end{tabular}
\end{table}

\section{Performance Analysis}
\label{sec:performance}

\subsection{Comparative Analysis}
\label{subsec:comparative}

The objective of this section is to evaluate numerically the
performance of the SA1 multiwavelet, and compare it to some well-known
orthogonal and biorthogonal multiwavelets, with the criteria being
coding gain (CG)~\cite{SV-93,T-05}, Sobolev smoothness
(S)~\cite{J-98}, symmetry/anti-symmetry (SA), and length. The CG for
orthogonal transforms is a good indication of the performance in
signal processing. It is the ratio of arithmetic and geometric means
of channel variances
\begin{equation}\label{eq:27}
  CG = \frac{\frac{1}{r} \sum_{i=1}^r \sigma_i^2}{\left( \prod_{i=1}^r
    \sigma_i^2 \right)^{1/r}}.
\end{equation}

Coding gain is one of the most important factors to be considered in
many applications. A transform with higher coding gain compacts more
energy into a smaller number of coefficients. A comparison of CGs is
given in table~\ref{tab:02}. Sobolev regularity, symmetry/antisymmetry
properties and the length of the given multiwavelets are also
included. The CG is computed using a first order Markov model AR(1)
with intersample autocorrelation coefficient
$\rho=0.95$~\cite{V-98}. Obviously, despite the short length (only two
matrix coefficients) of the simple multiwavelet SA1, its CG is
better than that of most of the orthogonal and biorthogonal
multiwavelets.

\begin{table}
  \caption{Comparison of CG, symmetric/antisymmetric properties (SA),
    Sobolev regularity (S) and autocorrelation matrix $R_x$ of AR(1)
    and $\rho = 0.95$.}
  \label{tab:02}
  \begin{tabular}{ccccc}
    \hline
    Multiwavelet & CG & S & SA & Length \\
    \hline
    Biorthogonal (dual to Hermitian cubic spline) ({\tt 'bih32s'}) \cite{AZC-07} & 1.01 & 2.5 & yes & 3/5 \\
    Biorthogonal Hermitian cubic spline ({\tt 'bih34n'}) \cite{S-98a} & 1.51 & 2.5 & yes & 3/5 \\
    Integer Haar \cite{CP-01} & 1.83 & 0.5 & yes & 2 \\
    CL \cite{CL-96} & 2.06 & 1.06 & yes & 3 \\
    SA1 & 2.13 & 0.5 & yes & 2 \\
    SA4 \cite{STT-00} & 3.73 & 0.99 & yes & 4 \\
    GHM \cite{GHM-94} & 4.41 & 1.5 & yes & 4 \\
    \hline
  \end{tabular}
\end{table}

\subsection{Pre- and Postfilters}
In the case of a scalar-valued input signal to the multiple-input
multiple-output (MIMO) filter bank it is necessary to vectorize the
input. In other words, a pre-processing step (called {\em
  prefiltering}) must be inserted at the beginning of the analysis
multifilter; in a symmetric way, a {\em postfilter} must be
incorporated at the end of the synthesis filter bank. In practical
applications, a popular choice is based on the Haar transforms
\cite{CMP-98,TST-99}.

Haar pre- and postfilters have the advantage of simultaneously
possessing symmetry and orthogonality, and no multiplication is needed
if one ignores the scaling factor. The prefilters should be as short
as possible to preserve the time localization of the wavelet
decomposition. The choice of prefilter is often critical for the
results, and should be made depending on the application at hand and
the type of signal to be processed.

The postfilter $\calN(z)$ that accompanies the prefilter $\calM(z)$
must satisfy $\calN(z) \calM(z) = I$. The matrix coefficients of the
pre- and postfilters we used are shown in table \ref{tab:03}.

\begin{table}
  \caption{The matrix coefficients of the pre- and postfilters for
    GHM, CL, and SA1 orthogonal multiwavelets}
  \label{tab:03}
  \begin{tabular}{ccc}
    \hline
    & $\calM(z)$ & $\calN(z)$ \\
    \hline
    GHM &  $\dfrac{1}{8\sqrt{2}} \left(
    \begin{bmatrix}
      3 & 10 \\
      0 & 0
    \end{bmatrix} + 
    \begin{bmatrix}
      3 & 0 \\
      8 \sqrt{2} & 0
    \end{bmatrix} z^{-1} \right)$ & $\dfrac{1}{10} \left(
    \begin{bmatrix}
      0 & 10 \\
      0 & -3
    \end{bmatrix} z + 
    \begin{bmatrix}
      0 & 0 \\
      8 \sqrt{2} & -3
    \end{bmatrix} \right)$ \\[10pt]
    CL & $
    \begin{bmatrix}
      1/4 & 1/4 \\
      1/(1+\sqrt{7}) & -1/(1+\sqrt{7})
    \end{bmatrix}$ & $
    \begin{bmatrix}
      2 & (1+\sqrt{7})/2 \\
      2 & -(1+\sqrt{7})/2
    \end{bmatrix}$ \\[10pt]
    SA1 & $ \dfrac{1}{\sqrt{2}}
    \begin{bmatrix}
      1 & 1 \\
      1 & -1
    \end{bmatrix}$ & 
    $\dfrac{1}{\sqrt{2}}
    \begin{bmatrix}
      1 & 1 \\
      1 & -1
    \end{bmatrix}$ \\
    \hline
  \end{tabular}
\end{table}

\subsection{Dyadic approximation of a 1D signal}

As explained in section~\ref{subsec:dyadic}, the customer often
prefers a simple dyadic approximation of the coefficients.  For a
chosen $b_0$ and $J$, we let $\hat\bolds_{1,J}$ be the signal
obtained by decomposing and reconstructing through $J$ levels, using
$(b_0+1)$ bits of accuracy. We define the {\em quantization error} by
the norm
\begin{equation*}
  \left\| \epsilon_{1,J} \right\|_\infty = \max_i \left| \epsilon_{1,J}(i) \right|,
\end{equation*}
where
\begin{equation*}
  \epsilon_{1,J}(i) = \bolds(i) - \hat \bolds_{1,J}(i).
\end{equation*}

We mainly will consider balanced and non-balanced multiwavelet
decomposition and synthesis without processing with a 2-bit
approximation of the coefficient $\sqrt{3}$, i.~e.~$b_0=1$, $\sqrt{3}
\approx 3/2$.

The non-balanced DMWT converts the scalar input signal into a
vector input signal by simple partitioning. The balanced DWMT uses a
preprocessing and postprocessing step.

\begin{figure}
  \includegraphics[height=2.7in]{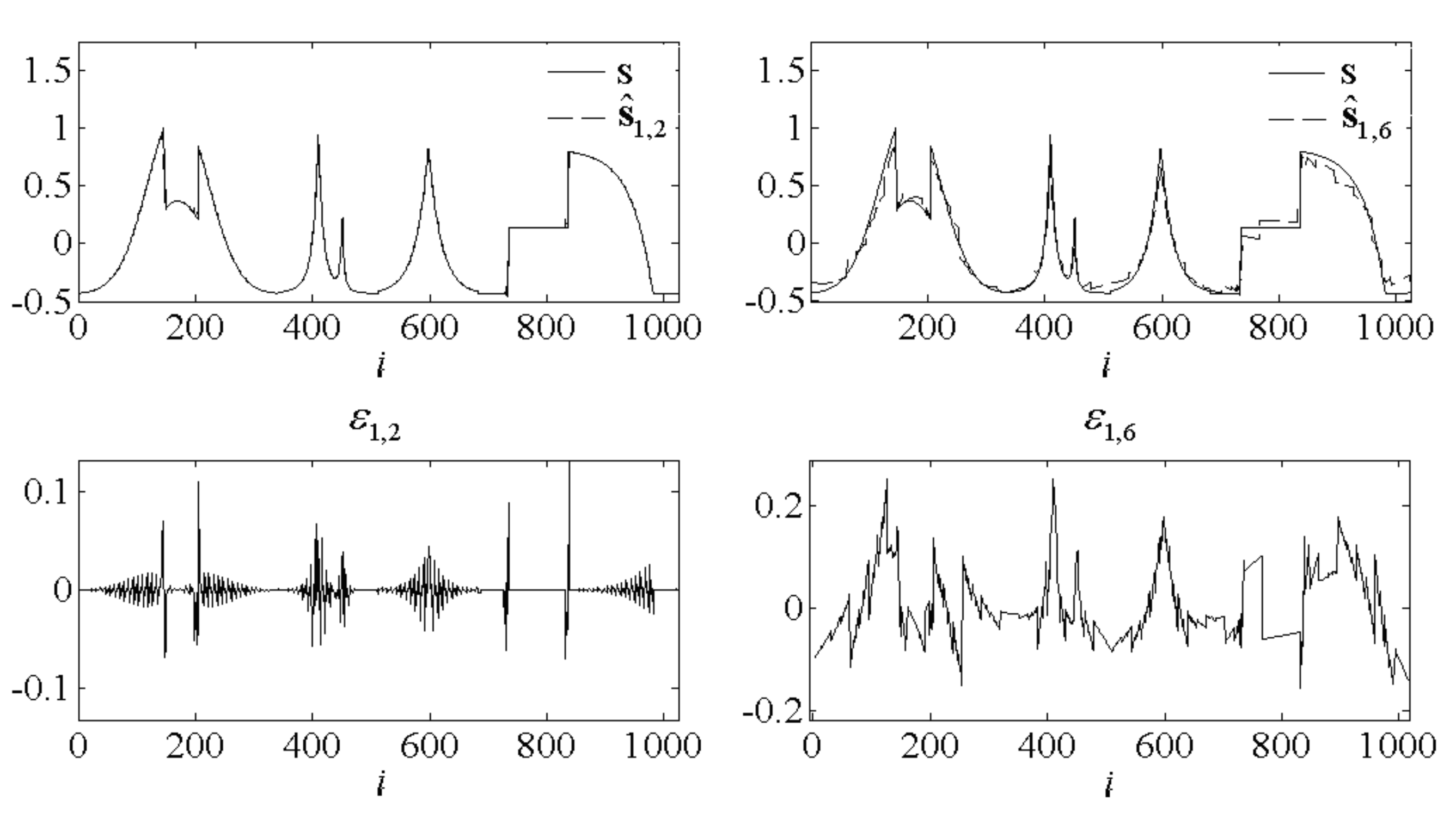} \\
  (a) \hspace{2.3in} (b)
  \caption{Influence of the number of decomposition levels $J$ of the
    2-bit dyadic approximations for the coefficient $\sqrt{3}$ (
    i.~e. $\sqrt{3} \approx 3/2$) of the balanced SA1 multiwavelet
    analysis and synthesis filter bank without processing, for the
    {\tt 'Piece-Regular'} signal with $i = 2^{10}$ samples through (a)
    $J = 2$ levels with $\|\epsilon_{1,2}\|_\infty = 0.1317$; (b) $J =
    6$ levels with $\|\epsilon_{1,6}\|_\infty = 0.2535$. In both (a)
    and (b), the top image shows the original signal $\bolds$ and
    reconstruction $\hat \bolds_{1,J}$; the bottom image shows the
    error $\epsilon_{1,J}$.}
  \label{fig:06}
\end{figure}

Results of using the SA1 balanced filter bank with the Haar pre- and
postfilters given in table~\ref{tab:03}, for the {\tt 'Piece-Regular'}
signal with $2^{10}$ samples, through 2 and 6 decomposition levels,
with their these quantization errors are shown in
fig.~\ref{fig:06}(a), (b). Obviously, increasing the decomposition
level leads to an increase in quantization error.

It is important to note that the reconstructed signal follows the
input signal, except at spikes where the local maxima of the
quantization errors are achieved. The best results are obtained at
small decomposition levels, between 1 and 4, as shown in
fig.~\ref{fig:06}(a). Higher decomposition levels lead to an increase
of the quantization errors over the input signal; see
fig.~\ref{fig:06}(b).

\begin{figure}
  \includegraphics[height=2.7in]{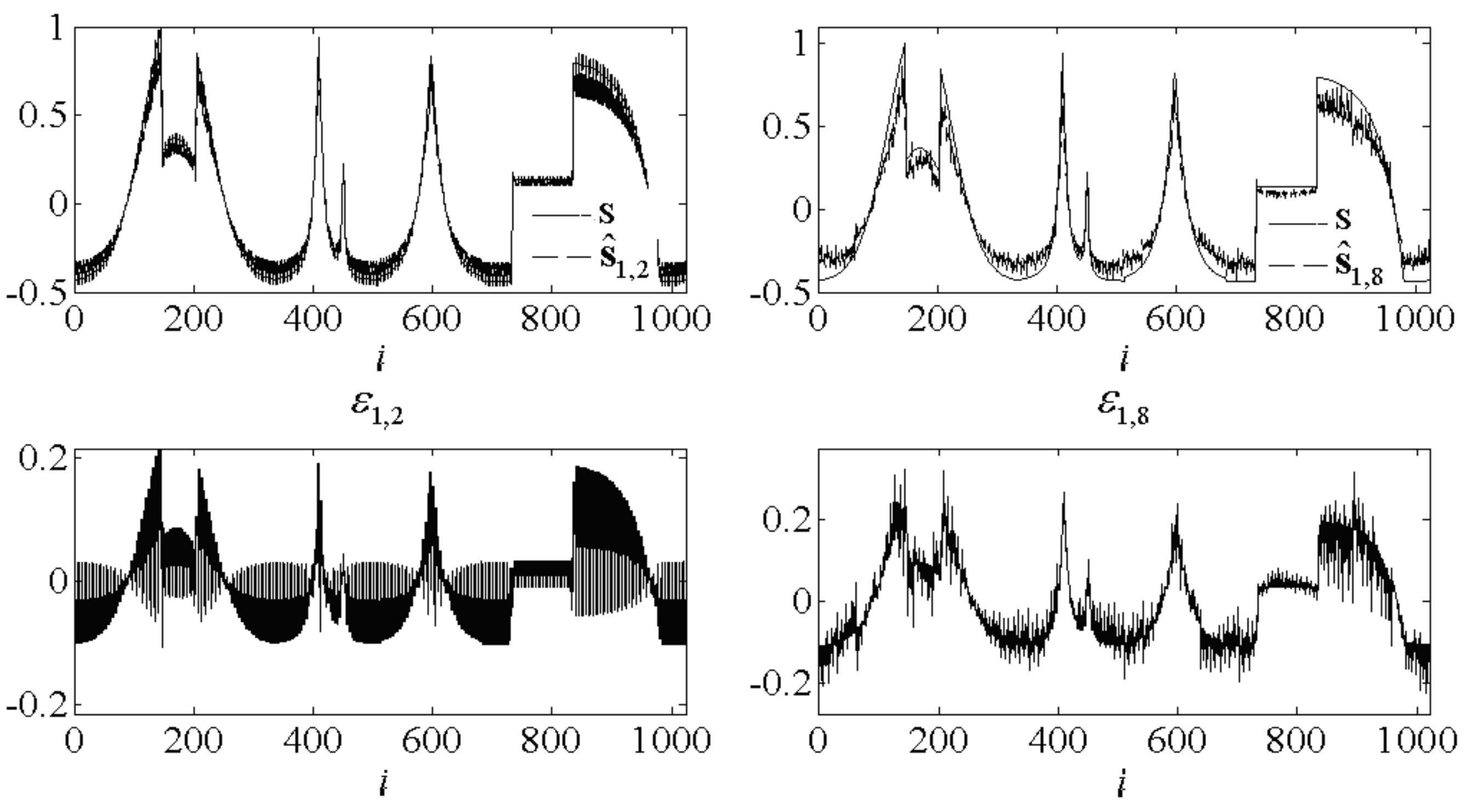} \\
  (a) \hspace{2.3in} (b)
  \caption{Influence of the number of decomposition levels $J$ of the
    2-bit dyadic approximations for the coefficient $\sqrt{3}$ (
    i.~e. $\sqrt{3} \approx 3/2$) of the non-balanced SA1 multiwavelet
    analysis and synthesis filter bank without processing, for the
    {\tt 'Piece-Regular'} signal with $i = 2^{10}$ samples through (a)
    $J = 2$ levels with $\|\epsilon_{1,2}\|_\infty = 0.2148$; (b) $J =
    8$ levels with $\|\epsilon_{1,8}\|_\infty = 0.3239$. In both (a)
    and (b), the top image shows the original signal $\bolds$ and
    reconstruction $\hat\bolds_{1,J}$; the bottom image shows the
    error $\epsilon_{1,J}$.}
  \label{fig:07}
\end{figure}

In contrast to the balanced SA1 multiwavelet, the quantization errors
of the non-balanced SA1 multiwavelet are very different, and poor. The
reconstructed signal oscillates around the analyzing signal, and the
quantization errors are bigger than the balanced version, even at low
levels, as shown in fig.~\ref{fig:07}(a).

\begin{table}
  \caption{Dependence of the norm $\|\epsilon_{1,J}\|_\infty$ on the
    number of decomposition levels $J$ for the balanced SA1 multiwavelet without
    processing. The 2-bit dyadic approximation of the coefficient
    $\sqrt{3}$ (i.~e.~$\sqrt{3} \approx 3/2$) is applied to the {\tt
      'Piece-Regular'} signal.}
  \label{tab:04}
  \begin{tabular}{cccccc}
    \hline
    Decomp. & \multicolumn{5}{c}{Test signal length} \\
    levels & $2^9$ samples & $2^{10}$ samples & $2^{11}$ samples &
    $2^{12}$ samples & $2^{13}$ samples \\
    \hline
    1  & {\bf 0.06084} & {\bf 0.06149} & {\bf 0.06695} & {\bf 0.06646}
    & {\bf 0.06103} \\
    2  & 0.12786 & 0.13167 & 0.12747 & 0.14190 & 0.10010 \\
    3  & 0.18903 & 0.13848 & 0.13056 & 0.14656 & 0.13211 \\
    4  & 0.23077 & 0.18432 & 0.18191 & 0.17597 & 0.13931 \\
    5  & 0.26208 & 0.21887 & 0.16843 & 0.20945 & 0.14795 \\
    6  & {\bf 0.29277} & 0.25346 & 0.21485 & 0.17841 & 0.18582 \\
    7  & 0.26932 & {\bf 0.28641} & 0.26010 & 0.22375 & 0.17793 \\
    8  &         & 0.26101 & 0.28482 & 0.26315 & 0.22557 \\
    9  &         &         & {\bf 0.29099} & 0.29428 & 0.26304 \\
    10 &         &         &         & {\bf 0.30911} & {\bf 0.29569} \\
    11 &         &         &         &         & 0.28781 \\
    \hline    
  \end{tabular}
\end{table}

\begin{table}
  \caption{Dependence of the norm $\|\epsilon_{1,J}\|_\infty$ on the
    number of decomposition levels $J$ for the balanced SA1
    multiwavelet without processing. The 2-bit dyadic approximation of
    the coefficient $\sqrt{3}$ (i.~e.~$\sqrt{3} \approx 3/2$) is
    applied to the {\tt 'Piece-Polynomial'} signal.}
  \label{tab:05}
  \begin{tabular}{cccccc}
    \hline
    Decomp. & \multicolumn{5}{c}{Test signal length} \\
    levels & $2^9$ samples & $2^{10}$ samples & $2^{11}$ samples &
    $2^{12}$ samples & $2^{13}$ samples \\
    \hline
    1  & {\bf 0.05834} & {\bf 0.13124} & {\bf 0.13118} & {\bf 0.13073} & {\bf 0.05770} \\
    2  & 0.13597 & 0.25323 & 0.27918 & 0.27996 & 0.13145 \\
    3  & 0.25061 & 0.26580 & 0.38819 & 0.29392 & 0.26348 \\
    4  & 0.25272 & {\bf 0.37841} & 0.32504 & {\bf 0.39067} & 0.28238 \\
    5  & {\bf 0.31074} & 0.32303 & {\bf 0.37016} & 0.31828 & {\bf 0.39400} \\
    6  & 0.26599 & 0.30530 & 0.33977 & 0.38265 & 0.33192 \\
    7  & 0.26648 & 0.36935 & 0.36768 & 0.33015 & 0.37298 \\
    8  &         & 0.36903 & 0.36846 & 0.35761 & 0.34802 \\
    9  &         &         & 0.36837 & 0.36052 & 0.37613 \\
    10 &         &         &         & 0.36055 & 0.37174 \\
    11 &         &         &         &         & 0.37168 \\
    \hline    
  \end{tabular}
\end{table}

\begin{figure}
  \includegraphics[width=2in]{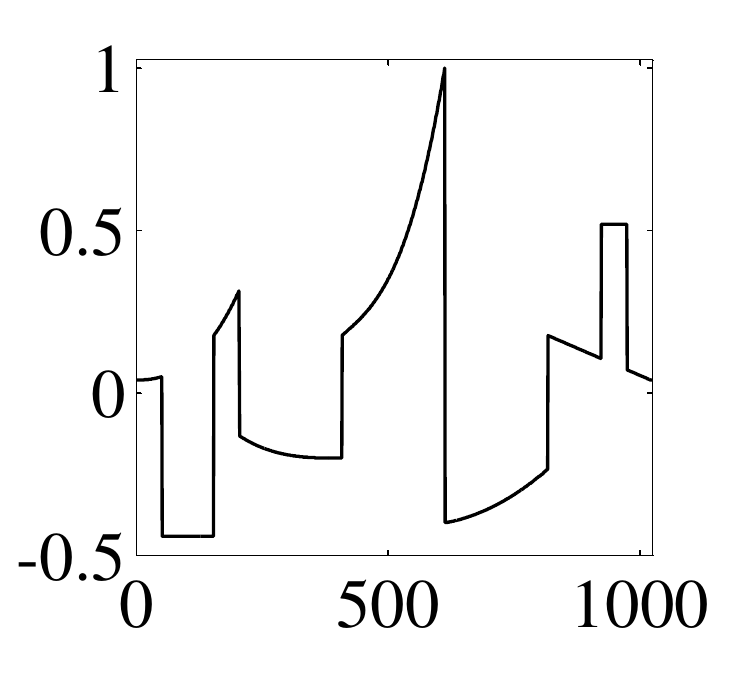}
  \caption{Additional test signal {\tt 'Piece-Polynomial'}}
  \label{fig:08}
\end{figure}

Detailed information about the minimal and maximal quantization
errors, measured by the norm $\| \epsilon_{1,J} \|_\infty$ for the
test signal {\tt 'Piece-Regular'} (see fig.~\ref{fig:06}), and the
additional test signal {\tt 'Piece-Polynomial'} (see
fig.~\ref{fig:08}) with $2^9$ to $2^{13}$ samples are tabulated in
tables \ref{tab:04} and \ref{tab:05}.

The minimal quantization errors are at the first decomposition level
for all lengths of both test signals, while the maximum errors depend
on the test signal. There are only minor differences in the minimal
quantization errors for the {\tt 'Piece-Regular'} signal for all
signal lengths, which means that the minimal quantization error is
almost independent of the length of this signal.

The minimal quantization errors for the {\tt 'Piece-Polynomial'}
signal for signal lengths $2^{10}$ to $2^{12}$ samples are nearly
equal, but for $2^9$ and $2^{13}$ samples decrease by more than a
factor of 2.

The maximum quantization errors for the {\tt 'Piece-Regular'} signal
are at the first and next to last levels and are nearly equal, and for
the {\tt 'Piece-Polynomial'} signal at decomposition levels 4 or 5.

Note that the minimal quantization errors are in the range
$[0.01,0.05]$, but the maximal quantization errors are in the range
$[0.3,0.4]$, which leads to visible artifacts in signal and image
processing.

\subsection{Dyadic approximations for 2D signals}

In this subsection, we consider the influence of 2-bit ($\sqrt{3}
\approx 3/2$) and 3-bit ($\sqrt{3} \approx 7/4$) dyadic approximation
applied to analysis and reconstruction filter bank without processing,
in the context of the balanced and non-balanced SA1 multiwavelet. We
will consider the gray scale image {\tt 'Lena'} of size $256 \times
256$ pixels.

\begin{figure}
  \includegraphics[width=4.9in]{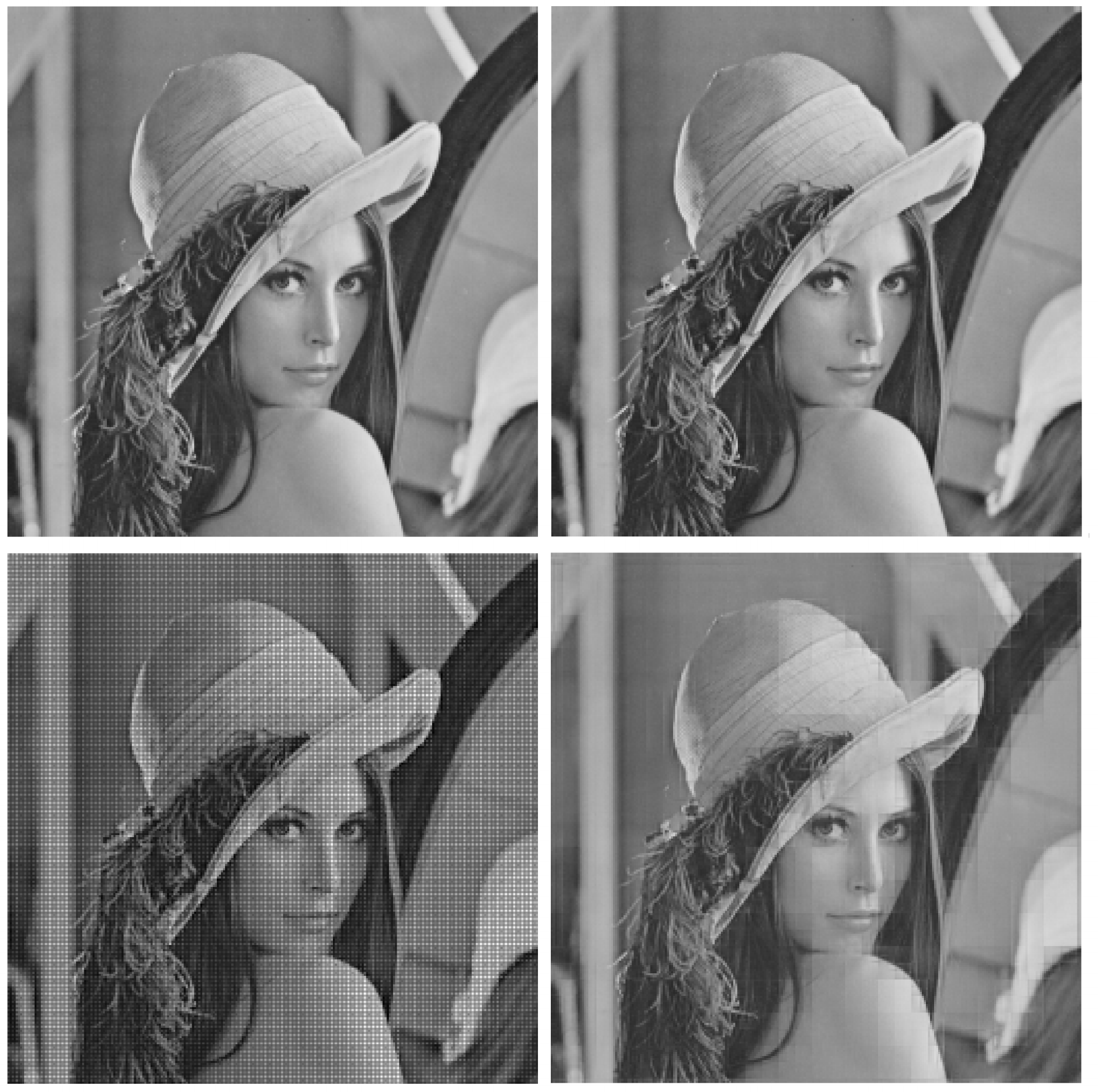} \\
  (a) \hspace{2.4in} (b)
  \caption{Influence of the number of decomposition levels for 2-bit
    dyadic approximations for the coefficient $\sqrt{3}$ (
    i.~e. $\sqrt{3} \approx 3/2$) of the SA1 multiwavelet analysis and
    synthesis filter bank without processing. (a) Non-balanced; top:
    $J=1$ level, PSNR = $\infty$; bottom: $J=4$ levels, PSNR = 17.12
    dB.  (b) Balanced; top: $J=1$ level, PSNR = 323.35 dB; bottom:
    $J=4$ levels, PSNR = 30.07 dB.}
  \label{fig:09}
\end{figure}

The first example illustrates the effect of the 2-bit dyadic
approximation of the coefficient $\sqrt{3}$ (i.~e.~$\sqrt{3} \approx
3/2$). As we can see see from fig.~\ref{fig:09}(a), image processing
with the SA1 non-balanced multiwavelet filter bank leads a grid
artifact over the whole image. Furthermore, with more than one
decomposition level the obtained images are usually unusable. It is
also important to note that after more than $J=4$ decomposition
levels, the multiwavelet processing with the balanced filter bank
shows visible square edge artifacts, even though the PSNR can be
high. This is shown in fig.~\ref{fig:09}(b).

\begin{figure}
  \includegraphics[width=4.9in]{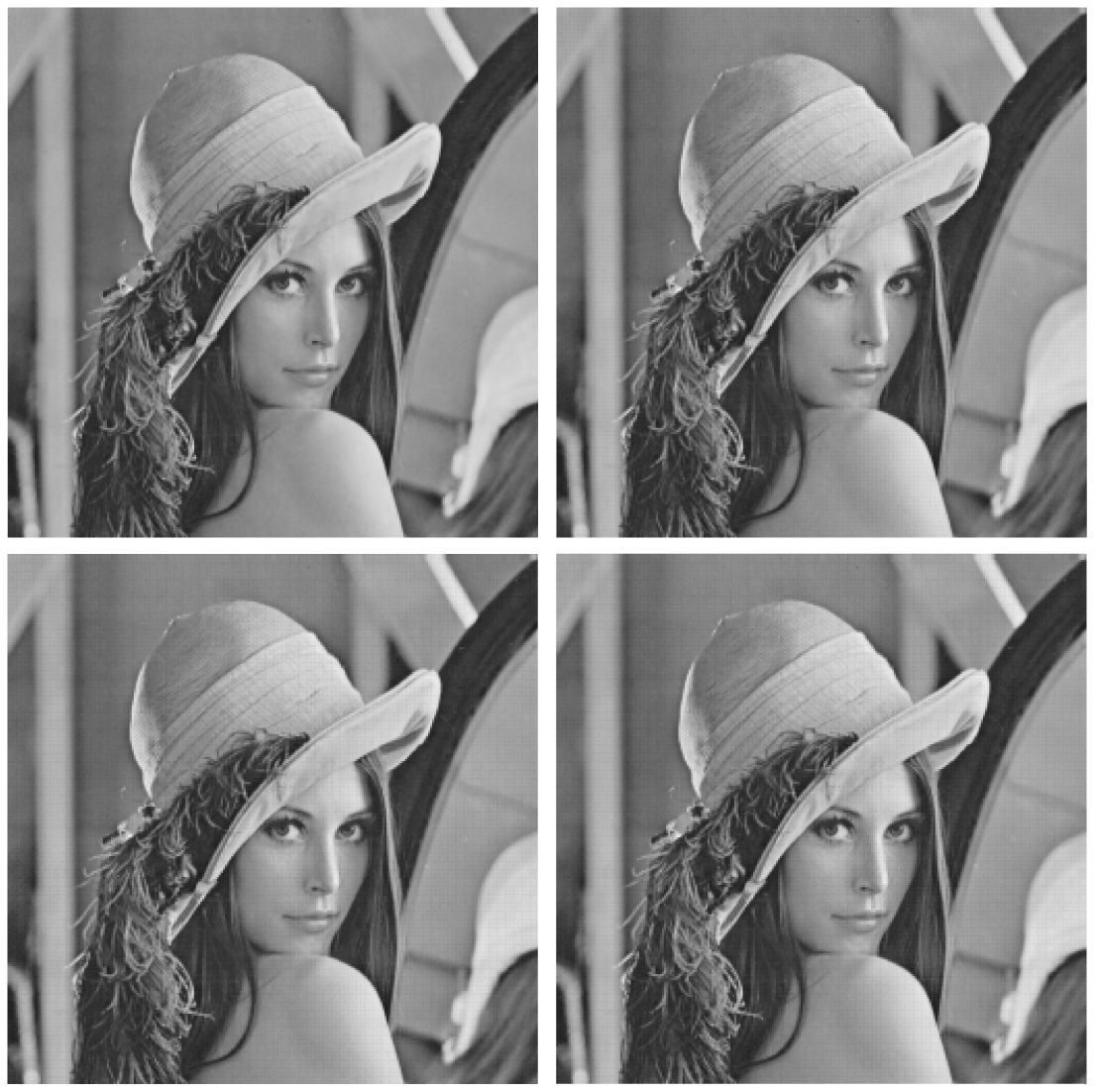}
  (a) \hspace{2.4in} (b)
  \caption{Influence of the number of decomposition levels for 3-bit
    dyadic approximations for the coefficient $\sqrt{3}$ (
    i.~e. $\sqrt{3} \approx 7/4$) of the SA1 multiwavelet analysis and
    synthesis filter bank without processing. (a) Non-balanced; top:
    $J=1$ level, PSNR = $\infty$; bottom: $J=4$ levels, PSNR = 38.07
    dB.  (b) Balanced; top: $J=1$ level, PSNR = 323.35 dB; bottom:
    $J=4$ levels, PSNR = 50.44 dB.}
  \label{fig:10}
\end{figure}

The second example illustrates a comparison of 3-bit dyadic
approximation of the coefficient $\sqrt{3}$ (i.e. $\sqrt{3} \approx
7/4$). It is obvious from fig.~\ref{fig:10}(a) and (b) that the SA1
multiwavelet processing with both non-balanced and balanced filter
bank leads to high quality images. Therefore, in some applications we
prefer 3-bit dyadic approximation of the coefficient $\sqrt{3}$, which
does not lead to undesired image artifacts.

\subsection{Image Denoising}
\label{subsec:imagedenoising}

In order to suppress the noise coefficients, the noisy image is first
transformed to the multiwavelet domain, and then soft or hard vector
thresholding is applied at various resolution levels.  This is
described in more detail in subsection \ref{subsec:denoising}.

\begin{figure}
  \includegraphics[width=1.5in]{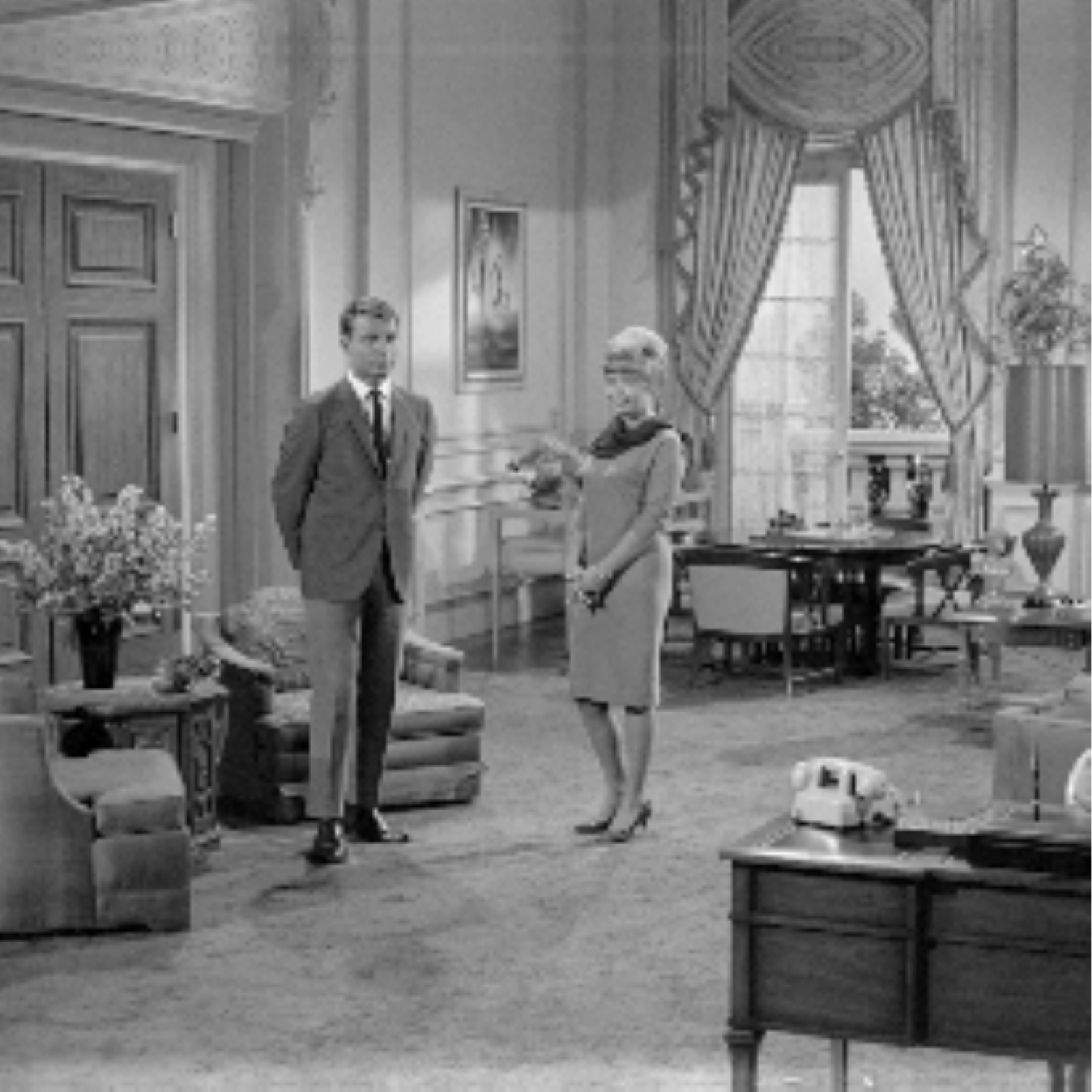} \quad
  \includegraphics[width=1.5in]{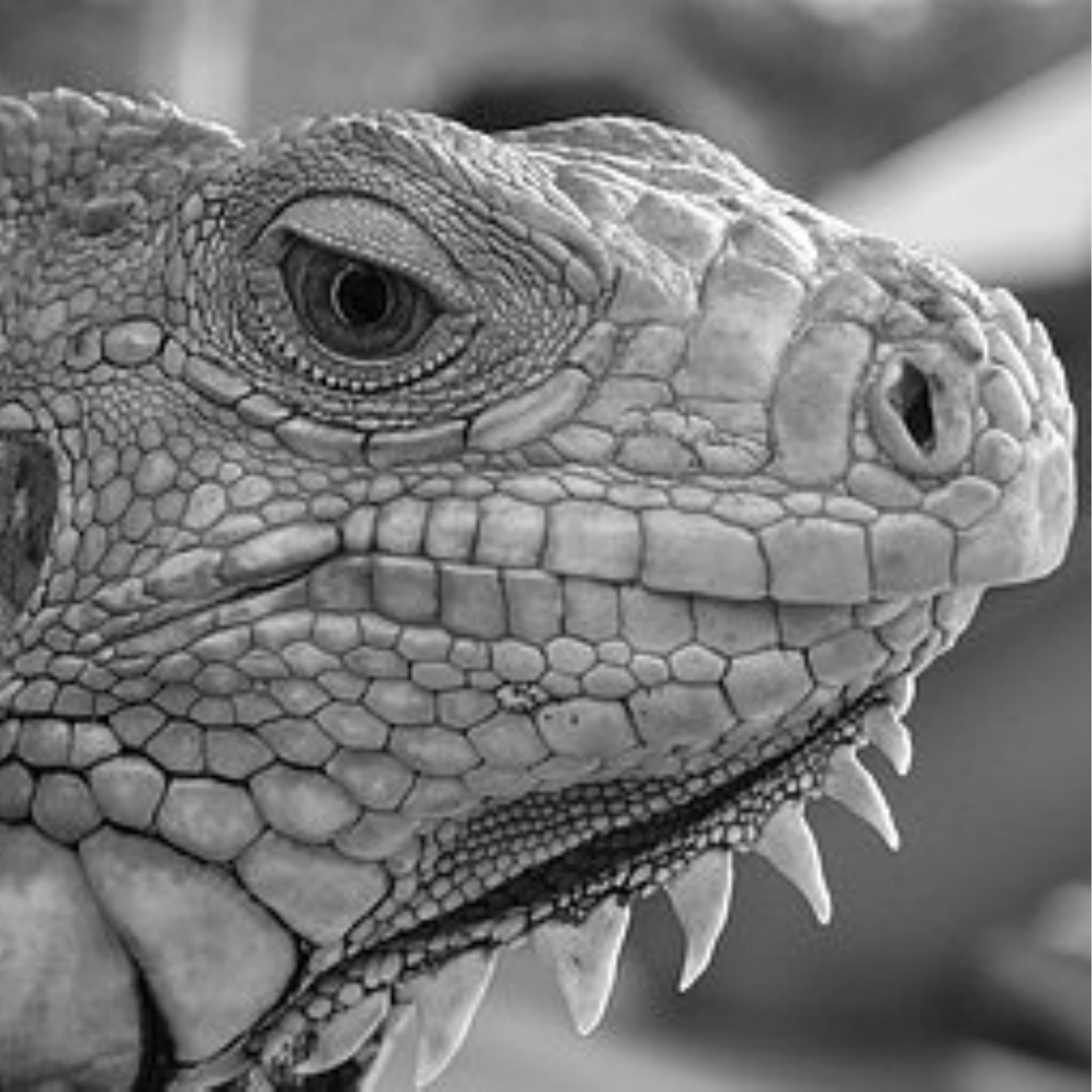} \quad
  \includegraphics[width=1.5in]{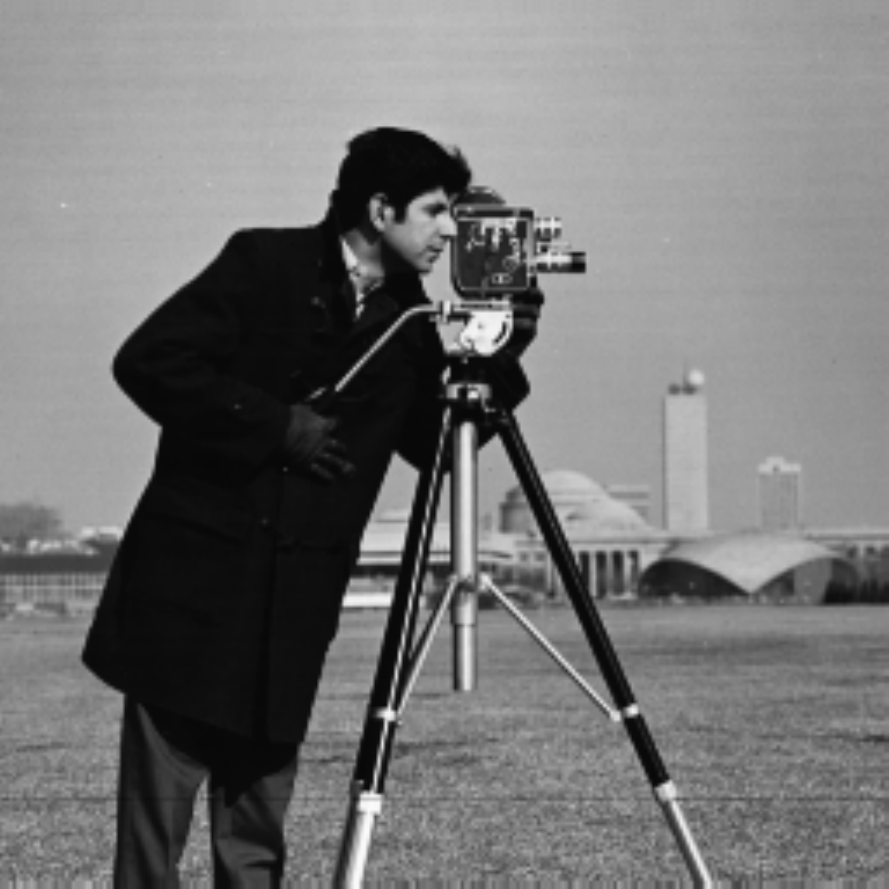} \\
  \caption{Test images of size $256 \times 256$ pixels ({\tt
      'Couple'}, {\tt 'Lizard'}, {\tt 'Cameraman'}). A $256 \times
    256$ version of {\tt 'House'} was also used (see fig.~\ref{fig:12}).} 
  \label{fig:11}
\end{figure}

\begin{figure}
  \includegraphics[width=1.5in]{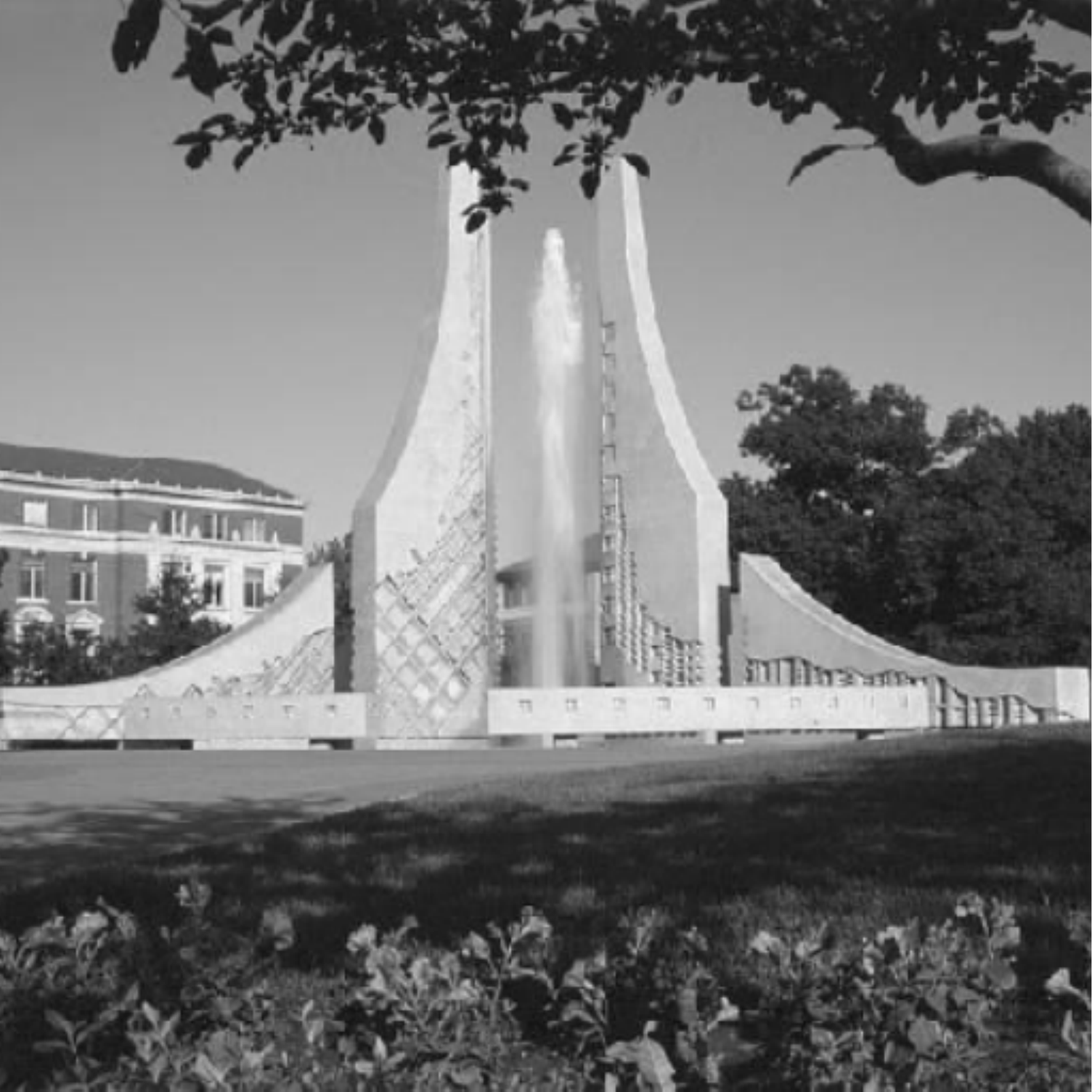} \quad
  \includegraphics[width=1.5in]{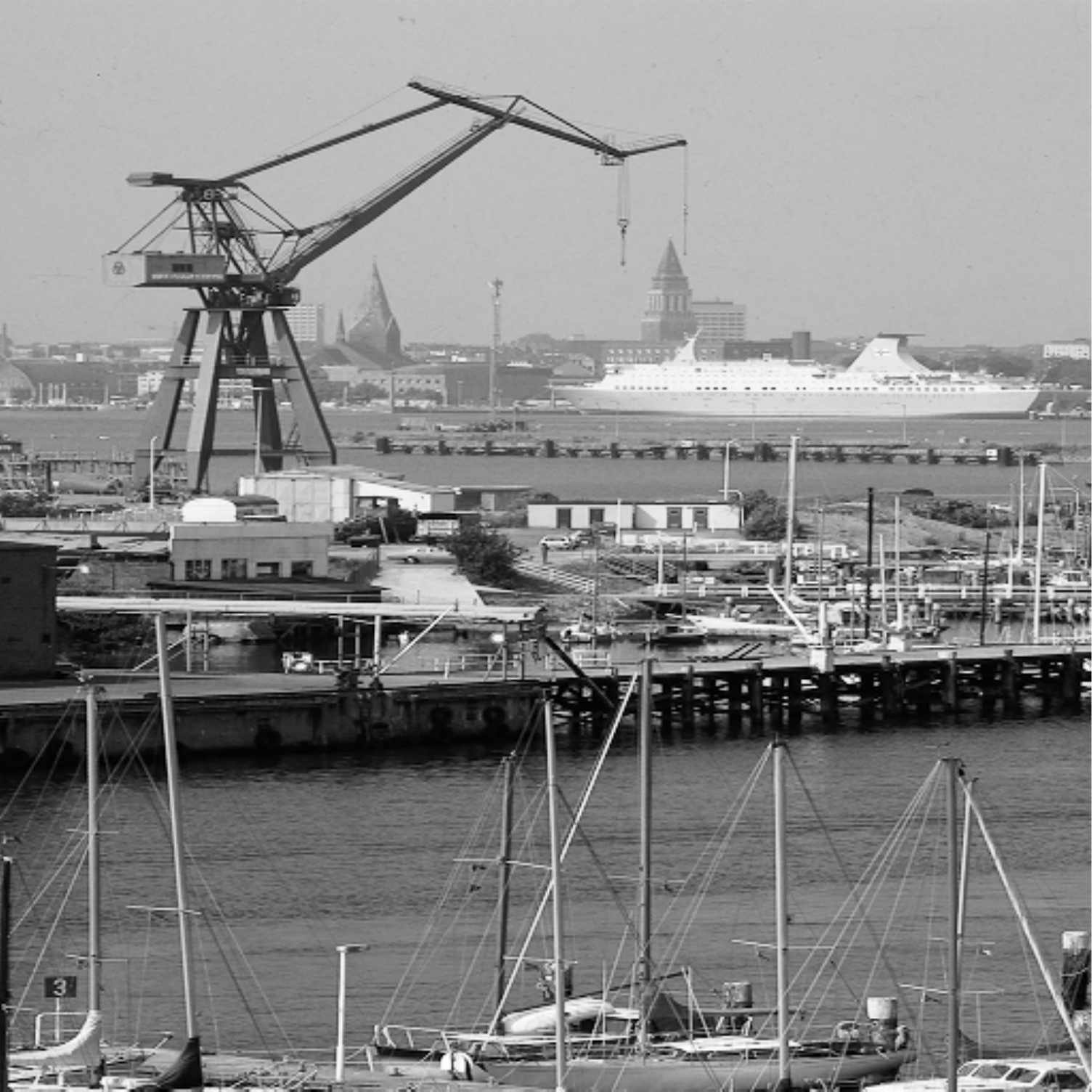} \quad
  \includegraphics[width=1.5in]{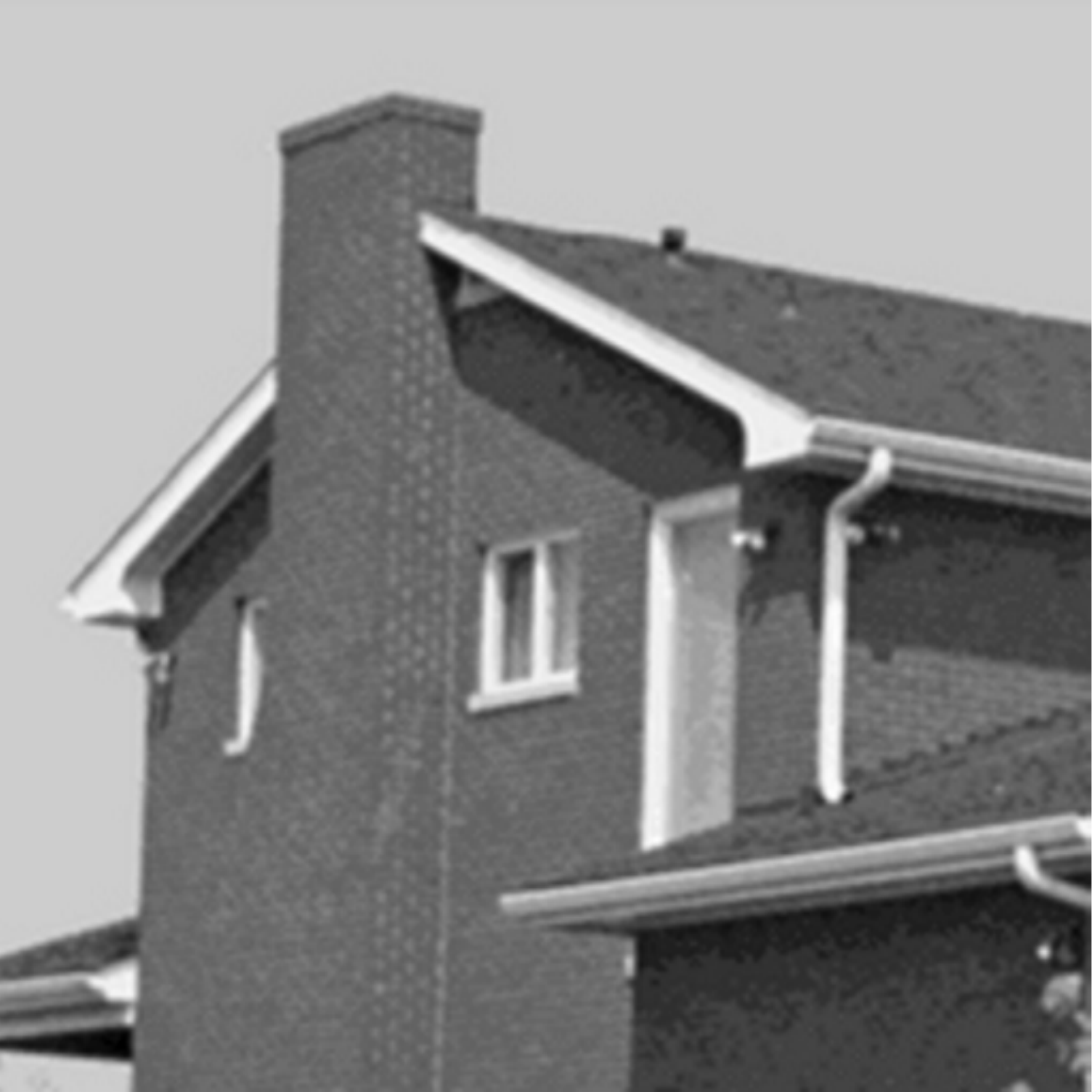} \\
  \caption{Test images of size $512 \times 512$ pixels ({\tt 'Fountain'},
    {\tt 'Kiel'}, {\tt 'House'}).}
  \label{fig:12}
\end{figure}

We apply image denoising to the test images shown in fig.~\ref{fig:11}
and fig.~\ref{fig:12}, using the lifting scheme
(eqs.~\eqref{eq:lifting} and~\eqref{eq:invlifting}) and the
multiwavelet decomposition in \cite{TSLT-00}, for three multiwavelets
GHM, CL, and SA1. The images are of size $256 \times 256$ or $512
\times 512$ pixels, with additive white Gaussian noise (AWGN) with
variance $\sigma = 10$. Either soft or hard vector thresholding is
applied.

The quality of an image denoising algorithm is determined by a
distortion measurement for determining how much information has been
lost when the reconstructed image is produced from the denoised
images. The most often used measurement is the {\em Mean Square Error}
(MSE). For an image of size $N \times N$, it is
\begin{equation*}
  \mbox{MSE} = \frac{1}{N^2} \sum_{x,y=1}^N \left( s(x,y) - \hat
  s(x,y) \right)^2,
\end{equation*}
where $s(x,y)$ is the original image, and $\hat s(x,y)$ is the denoised
image.

The {\em Peak Signal-to-Noise Ratio} (PSNR) is the MSE in decibels on
a logarithmic scale:
\begin{equation*}
  \mbox{PNSR} = 10 \log_{10}\left( 255^2/\mbox{MSE} \right) \mbox{ dB}.
\end{equation*}

The results shown in tables \ref{tab:06} and \ref{tab:07} provide a
comparison of PSNRs (in dB) for AWGN with variance $\sigma = 10$.
Best results are shown in boldface. Table \ref{tab:06} shows the
results for the images of size $256 \times 256$ pixels; table
\ref{tab:07} shows corresponding results for the images of size $512
\times 512$ pixels.

In general, the best performance for hard thresholding is provided by
SA1, in both tables; for soft thresholding, it is GHM.

\begin{table}
  \caption{Comparative results of PSNRs of denoised test images for
    three non-balanced orthogonal multiwavelets (GHM, CL, SA1) through
    $J=5$ decomposition levels with AWGN (with variance $\sigma = 10$)
    and vector thresholds. The test images are of size $256 \times
    256$ pixels.}
  \label{tab:06}
  \rotatebox{270}{
    \begin{tabular}{|ccccccccccccc|}
      \hline
      \multirow{4}{*}{Image} &
      \multirow{4}{*}{Threshold} &
      \multirow{4}{*}{Multiwavelet} &
      \multicolumn{10}{c|}{Decomposition Levels} \\
      \cline{4-13} 
      & & &
      \multicolumn{2}{c}{1} &
      \multicolumn{2}{c}{2} &
      \multicolumn{2}{c}{3} &
      \multicolumn{2}{c}{4} &
      \multicolumn{2}{c|}{5} \\
      \cline{4-13} 
      & & & \multicolumn{10}{c|}{balanced} \\
      \cline{4-13} 
      & & & no & yes & no & yes & no & yes & no & yes & no & yes \\
      \hline
      & & GHM & {\bf 25.79} & 28.48 & {\bf 24.96} & {\bf 28.12} & {\bf
        23.64} & {\bf 27.83} & {\bf 24.51} & {\bf 27.71} & {\bf 24.47} & {\bf 27.67} \\
      & soft & CL & 24.49 & 28.52 & 23.26 & 28.00 & 22.98 & 27.71 & 22.91 & 27.61 & 22.89 & 27.57 \\
      & & SA1 & 24.20 & {\bf 28.59} & 23.00 & 28.04 & 22.70 & 27.80 & 22.63 & 27.70 & 22.61 & {\bf 27.67} \\
      Couple & & & & & & & & & & & & \\
      & & GHM & 28.11 & 28.48 & 27.88 & 29.36 & 27.83 & 29.34 & 27.81 & 29.34 & 27.81 & 29.34 \\
      & hard & CL & {\bf 28.13} & {\bf 29.42} & 28.07 & 29.53 & 28.01 & 29.52 & 28.00 & 29.52 & 28.00 & 29.52 \\
      & & SA1 & 28.12 & {\bf 29.42} & {\bf 28.09} & {\bf 29.54} & {\bf 28.03} & {\bf
        29.54} & {\bf 28.02} & {\bf 29.53} & {\bf 28.01} & {\bf 29.53} \\
      \hline
      & & GHM & {\bf 25.59} & 27.33 & {\bf 24.65} & {\bf 26.80} & {\bf 24.30} & 
          {\bf 26.55} & {\bf 24.19} & {\bf 26.46} & {\bf 24.15} & {\bf 26.43} \\
      & soft & CL & 22.46 & 27.29 & 23.23 & 26.64 & 22.94 & 26.38 & 22.86 & 26.30 & 22.84 & 26.27 \\
      & & SA1 & 24.18 & {\bf 27.35} & 22.98 & 26.76 & 22.68 & 26.50 & 22.60 & 26.41 & 22.58 & 26.38 \\
      Lizard & & & & & & & & & & & & \\
      & & GHM & {\bf 28.13} & 28.87 & 27.96 & 28.99 & 27.91 & 28.99 & 27.91 & 28.99 & 27.91 & 28.99 \\
      & hard & CL & {\bf 28.13} & 28.97 & 28.09 & 29.07 & 28.05 & 29.07 & 28.04 & 29.07 & 28.04 & 29.07 \\
      & & SA1 & {\bf 28.13} & {\bf 29.03} & {\bf 28.11} & {\bf 29.16} & {\bf 28.08} 
      & {\bf 29.16} & {\bf 28.07} & {\bf 29.16} & {\bf 28.07} & {\bf 29.16} \\
      \hline
      & & GHM & {\bf 25.77} & 31.17 & {\bf 25.17} & 31.72 & {\bf 24.99}
      & 31.48 & {\bf 24.89} & 31.32 & {\bf 24.86} & 31.25 \\
      & soft & CL & 24.42 & 31.31 & 23.18 & 31.64 & 22.91 & 31.43 & 22.84 & 31.27 & 22.82 & 31.38 \\
      & & SA1 & 24.12 & {\bf 31.40} & 22.90 & {\bf 31.96} & 22.61 & {\bf
        31.79} & 22.54 & {\bf 31.65} & 22.52 & {\bf 31.59} \\
      House & & & & & & & & & & & & \\
      & & GHM & 28.12 & 30.96 & 27.91 & 31.77 & 27.91 & 31.87 & 27.91 & 31.88 & 27.91 & 31.88 \\
      & hard & CL & {\bf 28.13} & 31.18 & 28.09 & 31.99 & 28.04 & 32.19 & 28.03 & 32.20 & 28.03 & 32.20 \\
      & & SA1 & {\bf 28.13} & {\bf 31.21} & {\bf 28.12} & {\bf 32.14} &
      {\bf 28.09} & {\bf 32.27} & {\bf 28.07} & {\bf 32.27} & {\bf 28.06} & {\bf 32.27} \\
      \hline
      & & GHM & {\bf 26.20} & 29.62 & {\bf 25.42} & 29.56 & {\bf 25.17}
      & 29.33 & {\bf 25.06} & 29.20 & {\bf 25.02} & 29.16 \\
      & soft & CL & 24.80 & 29.71 & 23.64 & 29.53 & 23.36 & 29.29 & 23.28 & 29.17 & 23.26 & 29.13 \\
      & & SA1 & 24.58 & {\bf 29.78} & 23.41 & {\bf 29.66} & 23.11 & {\bf
        29.45} & 23.02 & {\bf 29.34} & 23.00 & {\bf 29.30} \\
      Cameraman & & & & & & & & & & & & \\
      & & GHM & {\bf 28.31} & 30.15 & {\bf 28.11} & 30.62 & {\bf 28.08}
      & 30.70 & {\bf 28.07} & 30.70 & {\bf 28.07} & 30.70 \\
      & hard & CL & 28.02 & 30.36 & 28.00 & 30.88 & 27.95 & 30.97 & 27.95 & 30.97 & 27.95 & 30.97 \\
      & & SA1 & 28.00 & {\bf 30.39} & 27.98 & {\bf 30.94} & 27.95 & {\bf
        31.04} & 27.94 & {\bf 31.04} & 27.94 & {\bf 31.04} \\
      \hline
    \end{tabular}
  }
\end{table}

\begin{table}
  \caption{Comparative results of PSNRs of denoised test images for
    three non-balanced orthogonal multiwavelets (GHM, CL, SA1) through
    $J=5$ decomposition levels with AWGN (with variance $\sigma = 10$)
    and vector thresholds. The test images are of size $512 \times
    512$ pixels.}
  \label{tab:07}
  \rotatebox{270}{
    \begin{tabular}{|ccccccccccccc|}
      \hline
      \multirow{4}{*}{Image} &
      \multirow{4}{*}{Threshold} &
      \multirow{4}{*}{Multiwavelet} &
      \multicolumn{10}{c|}{Decomposition Levels} \\
      \cline{4-13} 
      & & &
      \multicolumn{2}{c}{1} &
      \multicolumn{2}{c}{2} &
      \multicolumn{2}{c}{3} &
      \multicolumn{2}{c}{4} &
      \multicolumn{2}{c|}{5} \\
      \cline{4-13} 
      & & & \multicolumn{10}{c|}{balanced} \\
      \cline{4-13} 
      & & & no & yes & no & yes & no & yes & no & yes & no & yes \\
      \hline
      & & GHM & {\bf 25.97} & 30.49 & {\bf 25.13} & {\bf 30.27} & {\bf 24.81} & 
          {\bf 29.83} & {\bf 24.69} & {\bf 29.66} & {\bf 24.65} & {\bf 29.61} \\
      & soft & CL & 24.25 & {\bf 30.63} & 23.02 & 30.10 & 22.74 & 29.68 & 22.66 & 29.52 & 22.64 & 29.48 \\
      & & SA1 & 23.95 & 30.54 & 22.73 & 30.08 & 22.42 & 29.69 & 22.34 & 29.55 & 22.32 & 29.50 \\
      Fountain & & & & & & & & & & & & \\
      & & GHM & 28.00 & 30.71 & 27.72 & 31.14 & 27.67 & 31.19 & 27.66 & 31.20 & 27.66 & 31.20 \\
      & hard & CL & {\bf 28.13} & {\bf 30.92} & 27.98 & {\bf 31.37} & 27.88 & {\bf
        31.42} & 27.86 & {\bf 31.43} & 27.86 & {\bf 31.43} \\
      & & SA1 & 28.12 & 30.87 & {\bf 29.00} & 31.32 & {\bf 27.95}
      & 31.37 & {\bf 27.93} & 31.38 & {\bf 27.93} & 31.38 \\
      \hline
      & & GHM & {\bf 25.38} & 27.92 & {\bf 24.45} & {\bf 27.53} & {\bf 24.18}
      & {\bf 27.30} & {\bf 24.07} & {\bf 27.21} & {\bf 24.04} & {\bf 27.18} \\
      & soft & CL & 24.18 & 27.92 & 22.89 & 27.45 & 22.60 & 27.22 & 22.52 & 27.13 & 22.49 & 27.11 \\
      & & SA1 & 23.89 & {\bf 27.96} & 22.63 & 27.52 & 22.32 & 27.29 & 22.24 & {\bf 27.21} & 22.22 & {\bf 27.18} \\
      Kiel  & & & & & & & & & & & & \\
      & & GHM & 28.09 & 29.37 & 27.78 & 29.58 & 27.72 & 29.61 & 27.71 & 29.61 & 27.71 & 29.61 \\
      & hard & CL & 28.12 & {\bf 29.49} & 28.05 & {\bf 29.73} & 27.97 & {\bf 29.77} & 27.95 & {\bf 29.77} & 27.96 & {\bf 29.77} \\
      & & SA1 & {\bf 28.11} & {\bf 29.49} & {\bf 28.07} & 29.72 &{\bf
        28.02} & 29.76 & {\bf 28.00} & {\bf 29.77} & {\bf 28.00} & {\bf 29.77} \\
      \hline
      & & GHM & {\bf 25.62} & 33.29 & {\bf 25.09} & 34.14 & {\bf 24.93}
      & {\bf 34.63} & {\bf 24.84} & {\bf 34.30} & {\bf 24.81} & {\bf 34.17} \\
      & soft & CL & 24.12 & 33.42 & 22.83 & {\bf 34.94} & 22.56 & 34.44 & 22.49 & 34.13 & 22.47 & 34.01 \\
      & & SA1 & 23.80 & {\bf 33.44} & 22.55 & {\bf 34.94} & 22.25 & 34.51 & 22.18 & 34.25 & 22.16 & 34.15 \\
      House  & & & & & & & & & & & & \\
      & & GHM & {\bf 28.13} & 32.80 & 27.97 & 34.61 & 27.90 & 34.75 & 27.88 & 34.76 & 27.88 & 34.76 \\
      & hard & CL & {\bf 28.13} & {\bf 32.93} & 28.06 & {\bf 34.75} & 27.98 &
      {\bf 34.89} & 27.98 & {\bf 34.90} & 27.98 & {\bf 34.90} \\
      & & SA1 & {\bf 28.13} & 32.89 & {\bf 28.08} & 34.55 & {\bf 28.01}
      & 34.70 & {\bf 27.99} & 34.73 & {\bf 27.99} & 34.73 \\
      \hline
    \end{tabular}
  }
\end{table}

\section{Conclusion}
\label{sec:conclusion}

We have presented a novel mathematical algorithm for simple orthogonal
multiwavelet construction and a comprehensive design approach for
spectral matrix decomposition and efficient lifting scheme
implementation. We designed a simple orthogonal and regular
multiwavelet with symmetry properties. The matrix spectral
factorization algorithm of Youla and Kazanjian was used, effectively
increasing its precision. It can be used in the future to find new
exact supercompact multiwavelets with better regularity and CG. The
coding gain of the design is very good, especially considering its
simplicity. The obtained multiwavelet SA1 can be used in denoising
image processing as well other applications in communication
systems. By using the property that the inverse matrix of an integer
triangular matrix is also an integer triangular matrix, up to a scalar
factor, a lifting scheme based on a PLUS type matrix decomposition was
constructed. An important note for further work is that the lifting
scheme is a proper choice, as in the present work. We have shown that
the proposed lifting scheme with reduced accuracy of the present SA1
multiwavelet is also capable of denoising images at different
decomposition levels, and can be used for image denoising and
reversible integer to integer multiwavelet transforms.


\begin{acknowledgements}
  The authors would like to thank the anonymous reviewers for their
  careful reading and helpful suggestions.
\end{acknowledgements}

\bibliographystyle{spmpsci}      
\bibliography{sa1}   

\end{document}